\documentclass[10pt,twocolumn,letterpaper]{article}

\usepackage{wacv}
\usepackage{times}
\usepackage{epsfig}
\usepackage{graphicx}
\usepackage{amsmath}
\usepackage{amssymb}
\usepackage{booktabs}
\usepackage[makeroom]{cancel}
\usepackage{amsthm}
\usepackage{multirow}

\usepackage{makecell}
\usepackage{stfloats}
\usepackage{pifont}
\usepackage{caption}
\newcommand{\cmark}{\ding{51}}%
\newcommand{\xmark}{\ding{55}}%
\def\wacvPaperID{0645} 

\wacvalgorithmstrack   

\wacvfinalcopy 

\ifwacvfinal
\usepackage[breaklinks=true,bookmarks=false]{hyperref}
\else
\usepackage[pagebackref=true,breaklinks=true,colorlinks,bookmarks=false]{hyperref}
\fi

\begin{document}

\title{Line Search-Based Feature Transformation for Fast, Stable, and Tunable Content-Style Control in Photorealistic Style Transfer}

\author{Tai-Yin Chiu\\
University of Texas at Austin\\
{\tt\small chiu.taiyin@utexas.edu}
\and
Danna Gurari\\
University of Colorado Boulder\\
{\tt\small Danna.Gurari@colorado.edu}
}


\thispagestyle{empty}
\twocolumn[{%
\renewcommand\twocolumn[1][]{#1}%
\maketitle\vspace{-1.5em}
\begin{center}
    \centering
    \captionsetup{type=figure}
    \includegraphics[width=0.9\textwidth]{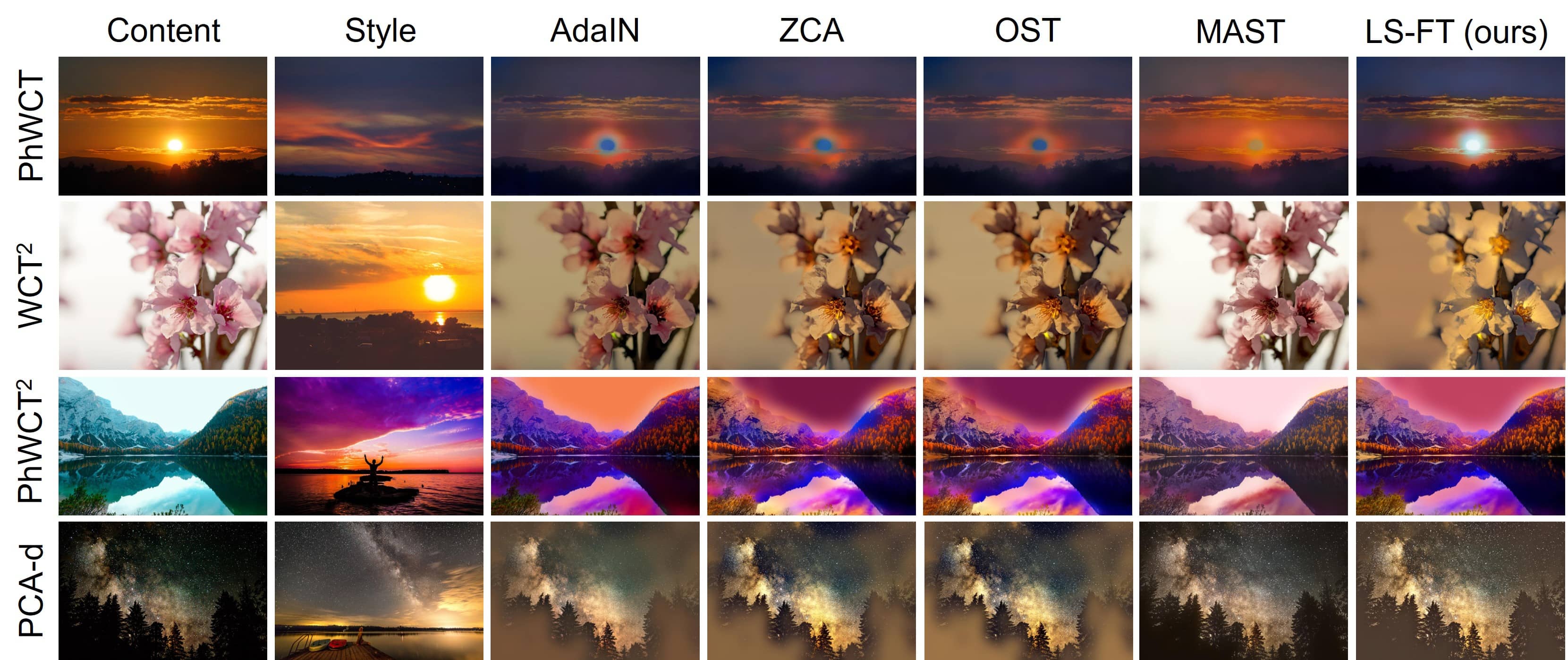}
    \captionof{figure}{\small Results are shown for embedding five transformations in four different autoencoder-based photorealistic style transfer models: WCT$^2$~\cite{yoo2019photorealistic}, PhotoWCT (PhWCT)~\cite{li2018closed}, PhotoWCT$^2$ (PhWCT$^2$)~\cite{chiu2021photowct2}, and a distilled model (PCA-d)~\cite{chiu2022pca}.  Our new transformation leads to a better balance between content preservation and style transfer than existing transformations.  Compared to AdaIN~\cite{huang2017arbitrary}, ZCA~\cite{li2017universal}, OST~\cite{lu2019closed}, and MAST~\cite{huo2021manifold}, our LS-FT can preserve better content with PhotoWCT and PCA-d, boost the stylization strength of WCT$^2$, and reach a better content-style balance with PhotoWCT$^2$.   }
    \label{fig:banner}
\end{center}%
}]

\begin{abstract}
   Photorealistic style transfer is the task of synthesizing a realistic-looking image when adapting the content from one image to appear in the style of another image.  Modern models commonly embed a transformation that fuses features describing the content image and style image and then decodes the resulting feature into a stylized image.  We introduce a general-purpose transformation that enables controlling the balance between how much content is preserved and the strength of the infused style.  We offer the first experiments that demonstrate the performance of existing transformations across different style transfer models, and demonstrate how our transformation performs better in its ability to simultaneously run fast, produce consistently reasonable results, and control the balance between content and style in different models. To support reproducing our method and models, we share the code at \textit{https://github.com/chiutaiyin/LS-FT}. 
\end{abstract}
\vspace{-1em}
\section{Introduction}
\label{sec:introduction}

Photorealistic style transfer is an image editing task that renders a content image with the style of another image, which we call the style image, such that the result looks like a realistic photograph to people.  In this paper, we tackle a key challenge of how to preserve the content while ensuring strong adoption of the style.  


   
\begin{figure}[!t]
    \centering
    \includegraphics[width=\linewidth]{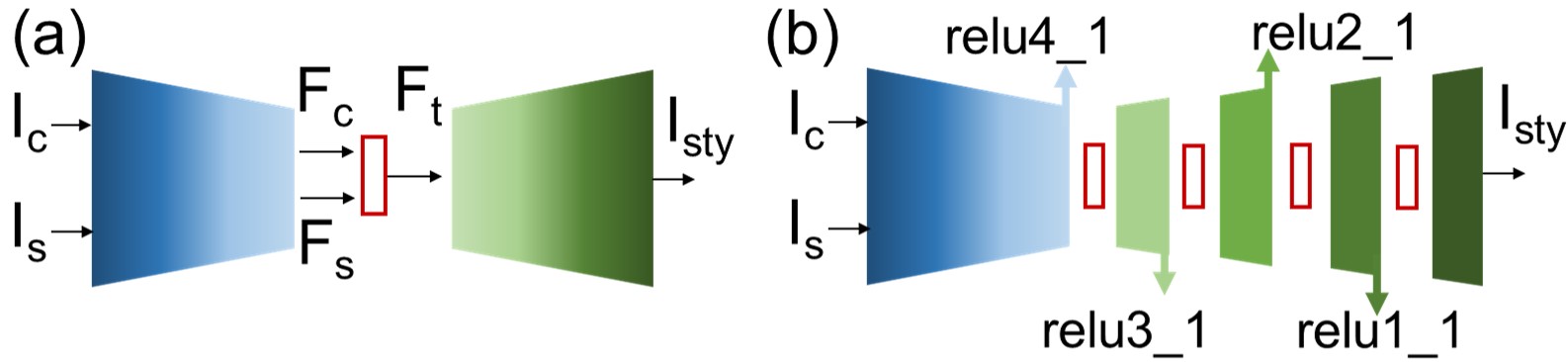}
    \caption{Autoencoder-based algorithms for photorealistic style transfer. (a) The basic model with a feature transformation (red box) at the bottleneck takes as input a content image $I_c$ and a style image $I_s$ and produces a stylized image $I_{sty}$. (b) PhotoWCT$^2$~\cite{chiu2021photowct2} with multiple transformations embedded in the decoder sequentially adapts the $\textit{relu4\_1}$ to the $\textit{relu1\_1}$ content features.}
    \label{fig:autoencoder_framework}
\end{figure}

The state-of-the-art approaches for photorealistic style transfer consist of an autoencoder with feature transformations~\cite{li2018closed,yoo2019photorealistic,an2020ultrafast,chiu2021photowct2,chiu2022pca}.  Such approaches are advantageous in that they can transfer style from any arbitrary style image (i.e., are universal), are fast since they predict in a single forward pass, and do not involve training on any style images (i.e., are style-learning-free).  The basic model contains an encoder to extract the features of the content and style images, a feature transformation to adapt the content feature with respect to the style feature, and a decoder to convert the adapted feature to the stylized image (exemplified in Fig.~\ref{fig:autoencoder_framework}(a)). More advanced models embed multiple transformations to achieve better aesthetic results (e.g., PhotoWCT$^2$~\cite{chiu2021photowct2} exemplified in Fig.~\ref{fig:autoencoder_framework}(b)).

A limitation of advanced models is the lack of a general-purpose, stable, one-size-fits-all transformation that yields content-style balance across all models, as exemplified qualitatively in Fig.~\ref{fig:banner}. The most commonly used transformations in photorealistic style transfer models are AdaIN~\cite{huang2017arbitrary} and ZCA~\cite{li2017universal}. Yet, both AdaIN and ZCA can fail to faithfully reflect the style of style images when embedded in WCT$^2$ and do not preserve content well when embedded in PhotoWCT (Fig.~\ref{fig:banner}). When embedded in PhotoWCT$^2$, AdaIN can suffer insufficient stylization strength while ZCA can introduce artifacts that ruin the photorealism. When embedded in PCA-d, ZCA and AdaIN can lead to severe artifacts, as will be shown in Sec.~\ref{sec:exp_content_style_balance}. 

Recently, an iterative feature transformation~\cite{chiu2020iterative} (IterFT) was proposed which has a control knob to tune between content preservation and style transferal and so can adaptively address the limitation of content-style imbalance across different models. However, as will be explained in Sec.~\ref{sec:preliminaries} and Sec.~\ref{sec:modified_IterFT}, this transformation in practice is unstable as it often produces poor results. Additionally, it is relatively slow.

In this work, we provide several contributions.  We expose the problem that existing transformations do not generalize well when used with different photorealistic style transfer models through extensive experiments.  We address the limitations of existing transformations by introducing a new transformation, which we call LS-FT.  Our experiments show that it consistently achieves a better balance of content preservation and stylization strength by permitting tuning between these two competing aims. Additionally, our experiments show that it runs 8x faster and consistently produces more reasonable results than the only other transformation that can tune between content preservation and stylization strength: IterFT. Ablation studies reveal the key mechanisms behind our transformation's improved performance: introduction of two steps to IterFT (centralization and decentralization) and line-search based optimization.  

\section{Related Works}
\label{sec:related_works}
\paragraph{Photorealistic style transfer models. }
DPST~\cite{luan2017deep} is the first deep neural network-based model for photorealistic style transfer. However, DPST is slow since it needs hundreds of times of feed-forward and back-propagation to render an image. To solve this issue, most modern photorealistic style transfer models use autoencoders to render an image in a single forward pass.\footnote{An exception is \cite{xia2020joint} which learns affine mappings to alter pixel values.}

Some of these autoencoder models~\cite{li2019learning,kim2022deep,cheng2021style,hong2021domain} are trained on style images.  Yet, these have not emerged as state-of-the-art for a variety of reasons.  For instance, some only work for low resolution input images, such as DTP~\cite{kim2022deep} which supports only 256$\times$256 resolution.  Others often produce unrealistic results, as exemplified for LST~\cite{li2019learning} and DSTN~\cite{hong2021domain} in the Supplementary Materials. 

The other models are style-learning-free autoencoders. To achieve stylization, they embed transformations, which are the focus of our work.  These models come with numerous advantages including the flexibility to embed different transformations to achieve different purposes, such as fast speed, strong stylization strength, and better photorealism. Style-learning-free models include PhotoWCT~\cite{li2018closed}, WCT$^2$~\cite{yoo2019photorealistic}, PhotoNAS~\cite{an2020ultrafast}, PhotoWCT$^2$~\cite{chiu2021photowct2}, and PCA-d~\cite{chiu2022pca}. However, a lingering challenge lies in knowing what transformations to embed in style transfer architectures.  For example, many popular models use ZCA as the feature transformation, yet prior work~\cite{chiu2021photowct2} has shown it leads to weaker style effects for WCT$^2$ and PhotoNAS.   

We introduce the first study where we pair popular transformations with multiple style transfer architectures.\footnote{PhotoNAS is excluded from the main paper since its model size is too large to even handle the small HD image resolution on the GPU.}  We demonstrate limitations of existing transformations and show that our new transformation, LS-FT, generalizes better due to its ability to balance content preservation and stylization strength while also running fast.

\vspace{-3.5mm}
\paragraph{Feature transforms for photorealistic style transfer. }
Recently, numerous transformations have been proposed.  AdaIN~\cite{huang2017arbitrary} is the simplest, which adapts the content feature to match the standard deviation vector and the mean vector of the style feature. Extending AdaIN, ZCA~\cite{li2017universal,chiu2019understanding} considers the cross correlation between channels and transforms the content feature to match the mean and  covariance of the style feature. Experimentally~\cite{li2017universal,li2018closed,an2020ultrafast}, ZCA results in stronger stylization strength than AdaIN. OST~\cite{lu2019closed} further modifies the covariance matching operation of ZCA to better preserve the content while maintaining stronger stylization strength. However, the improvement in content preservation is limited. A limitation across all these transformations is they lack a way to control the balance between content preservation and style transferal.

To address the limitations of prior transformations, iterative feature transformation (IterFT)~\cite{chiu2020iterative} was proposed.  Its main advantage over its prior work is that it supports tuning between content preservation and style transferal to meet the needs of each model. Yet, our experiments show this transformation often produces unreasonable results.  Additionally, it is relatively slow. To overcome these shortcomings of IterFT while taking advantage of its ability to be tuned to the needs of different models, we extend it and propose a new transformation dubbed LS-FT. Our experiments demonstrate that LS-FT not only realizes model adaptiveness in practice, but consistently produces reasonable results while performing 8x faster than IterFT. 

Of note, a recently proposed transformation MAST~\cite{huo2021manifold} takes a different approach from the aforementioned transformations and our LS-FT.  Specifically, while most transformations adapt the content feature as a whole, MAST adapts each content feature pixel with respect to the style feature pixels that are semantically close. Yet, as shown in Fig.~\ref{fig:banner}, MAST can only weakly stylize content images.   

We summarize the properties of each transformation in Tab.~\ref{tab:transformations}, demonstrated by our experiments.  Our transformation improves upon prior work by simultaneously being fast (e.g., faster or comparable speed to ZCA in Tab.~\ref{tab:speed}), rendering consistently good results (Fig.~\ref{fig:iter_vs_Moditer} and Fig. 9 in the Supplementary Materials), and achieving content-style control (Fig.~\ref{fig:content_style_losses} and Figs. 6-7 in the Supplementary Materials).

\begin{table}[t]
\setlength{\tabcolsep}{2.0pt}
    \small
    \centering
    \begin{tabular}{l c c c c c c}\toprule
             &  AdaIN & ZCA & OST & MAST & IterFT & LS-FT\\\midrule
        Fast & \cmark & \cmark & \cmark & \xmark & \xmark & \cmark \\
        Consistent Results & \cmark & \cmark & \cmark & \cmark & \xmark & \cmark \\
        Content-Style Control & \xmark & \xmark & \xmark & \xmark & \cmark & \cmark\\\bottomrule
    \end{tabular}
    \vspace{-1em}
    \caption{Comparison of our LS-FT to existing transformations.  Our LS-FT is the only one that simultaneously realizes three beneficial properties.}
    \label{tab:transformations}
\end{table}

\vspace{-3.5mm}
\paragraph{Image translation. }
Image translation~\cite{anokhin2020high,liang2021high,park2020swapping,park2020contrastive,isola2017image,liu2017unsupervised,richardson2021encoding,murez2018image} models are trained to render images from one domain in the style of another domain (ex: day $\to$ night) and so are able to adapt image style. Unlike photorealistic style transfer, image translation is not universal meaning that we need to retrain the models when the domains of interest change. Image translation goes beyond style transfer by also altering content (ex: straight hair $\to$ curly hair).

\section{Method}
We now introduce our feature transformation that we designed for general-purpose use across multiple style transfer architectures.  We begin by describing in Section~\ref{sec:preliminaries} the iterative feature transformation (IterFT) we redesign. Then we introduce the two key ingredients that enable the strengths of our transformation compared to IterFT: a modification that leads to consistently high quality results (Sec.~\ref{sec:modified_IterFT}) and a \emph{line search-based feature transformation} (LS-FT) which enables faster speeds (Sec.~\ref{sec:LS-FT}).

\subsection{Background - iterative feature transformation}
\label{sec:preliminaries}
Recall that IterFT is a feature transformation for style transfer that has been shown to support controlling the balance between content preservation and style transferal. It does so by making the final feature's second-order statistic resemble that of the style feature while maintaining its proximity to the content feature. 


Formally, following Fig.~\ref{fig:autoencoder_framework}(a), we let $\mathbf{F}_{t}$ be the feature transformed from the content feature $\mathbf{F}_c$ with reference to the style feature $\mathbf{F}_s$. For simplicity, a feature $\mathbf{F}$, which is a tensor of shape $C\times H\times W$ ($C$, $H$, $W$: channel length, height, width of $\mathbf{F}$) when produced from a neural network layer, is reshaped to a matrix of shape $C\times HW$. Iterative feature transformation (IterFT)~\cite{chiu2020iterative} makes the second-order statistic, Gram matrix $\frac{1}{H_cW_c}\mathbf{F}_t\mathbf{F}_t^\mathrm{T}$, of $\mathbf{F}_t$ close to that of $\mathbf{F}_s$ while maintaining the proximity of $\mathbf{F}_t$ to the content feature $\mathbf{F}_c$. Letting $n_c$ $=$ $H_c W_c$ and $n_s$ $=$ $H_s W_s$, IterFT solves the optimization problem in Eq.~\ref{eq:iterFT_obj} for $\mathbf{F}_t$ using gradient descent with the analytical gradient $\frac{\mathrm{d}l}{\mathrm{d}\mathbf{F}_t}$ in Eq.~\ref{eq:iterFT_grad}.
\begin{equation}
\small
    \min_{\mathbf{F}_t} l(\mathbf{F}_t) = \min_{\mathbf{F}_t} ||\mathbf{F}_t - \mathbf{F}_c||^2_2 + \lambda||\frac{1}{n_c}\mathbf{F}_t\mathbf{F}_t^\mathrm{T}-\frac{1}{n_s}\mathbf{F}_s\mathbf{F}_s^\mathrm{T}||^2_2,
    \label{eq:iterFT_obj}
\end{equation}
\begin{equation}
    \frac{\mathrm{d}l}{\mathrm{d}\mathbf{F}_t} = 2(\mathbf{F}_t - \mathbf{F}_c) + \frac{4\lambda}{n_c}(\frac{1}{n_c}\mathbf{F}_t\mathbf{F}_t^\mathrm{T}-\frac{1}{n_s}\mathbf{F}_s\mathbf{F}_s^\mathrm{T})\mathbf{F}_t,
    \label{eq:iterFT_grad}
\end{equation}
where $\lambda$ $>$ $0$ is the coefficient controlling the balance between content preservation and style transferal. With $\mathbf{F}_t$ initialized to $\mathbf{F}_c$, the final feature $\mathbf{F}_t$ is produced from $n_{upd}$ iterations of the update rule: $\mathbf{F}_t \leftarrow \mathbf{F}_t - \eta \frac{\mathrm{d}l}{\mathrm{d}\mathbf{F}_t}$,
where $\eta$ is the learning rate.

Advanced photorealistic style transfer models often embed multiple feature transformations.  This is exemplified in Fig.~\ref{fig:autoencoder_framework}(b) for PhotoWCT$^2$~\cite{chiu2021photowct2}, which has four IterFTs.  The first IterFT adapts the $\textit{relu4}\_1$ content feature with respect to the $\textit{relu4}\_1$ style feature for $n_{upd}$ iterations, followed by the second IterFT which adapts the $\textit{relu3}\_1$ content feature with respect to the $\textit{relu3}\_1$ style feature for $n_{upd}$ iterations until the last IterFT finishes adapting the $\textit{relu1}\_1$ content feature with respect to the $\textit{relu1}\_1$ style feature.

\subsection{Modified iterative feature transformation}
\label{sec:modified_IterFT}
\begin{figure}[t!]
    \centering
    \includegraphics[width=\linewidth]{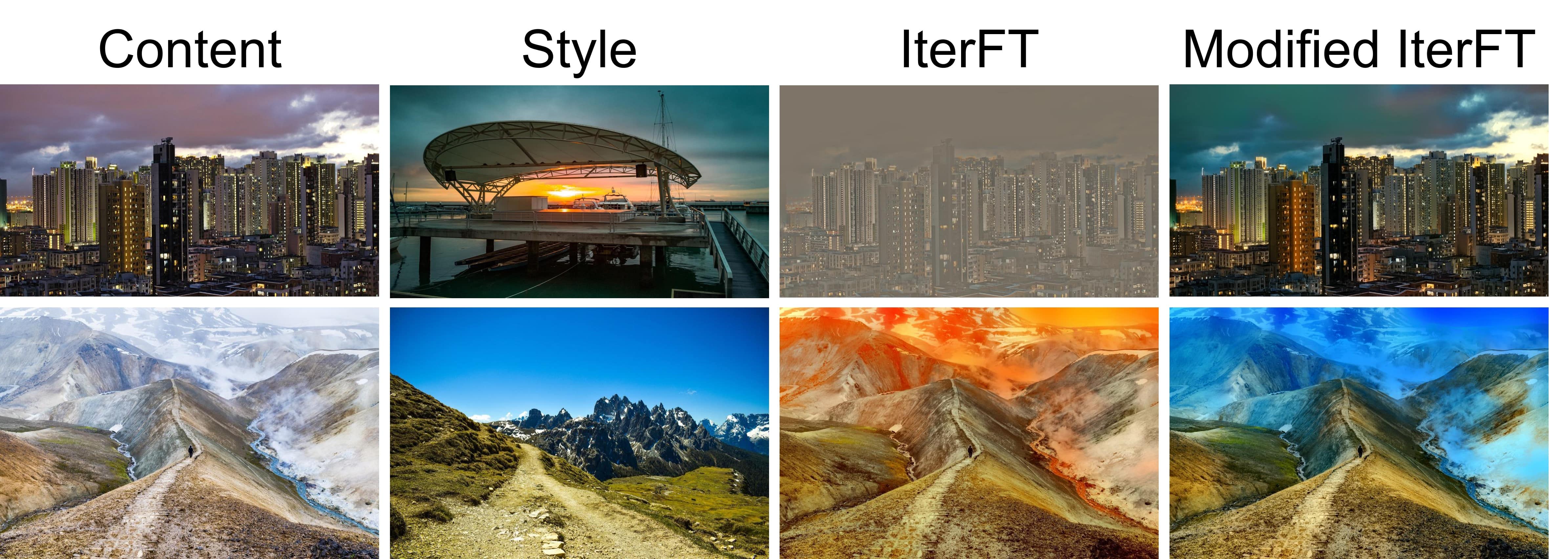}
    \caption{\small Qualitative results exemplify our modification to IterFT overcomes the instability of IterFT in producing reasonable stylized images.  These results are produced by embedding the transformations in the PhotoWCT$^2$ model.}
    \label{fig:iter_vs_Moditer}
\end{figure}

Our first modification is motivated by the observation that IterFT fails to stably produce reasonable results.  This is exemplified in Fig.~\ref{fig:iter_vs_Moditer}.  

We hypothesize that IterFT's failures stem from the fact that it relies on only one step (i.e., second-order statistic matching) rather than the three steps (i.e., centralization, second-order statistic matching, and decentralization) consistently employed by prior transformations~\cite{huang2017arbitrary,li2017universal,lu2019closed} that stably produce reasonable results.  Accordingly, we introduce centralization and decentralization before and after the second-order statistic matching. Our experimental results will validate the importance of these two steps by showing their addition to IterFT enables the resulting transformation to stably generate reasonable results (Sec.~\ref{sec:exp_content_style_balance}) and their removal from existing transformations (i.e., AdaIN~\cite{huang2017arbitrary} and ZCA~\cite{li2017universal}) lead to quality degradation in synthesized images (shown in the Supplementary Materials).  We suspect that the theory motivating the benefit of this modification is closely related to mean vector matching. While prior works~\cite{gatys2016image,li2017demystifying} focus on explaining the reason of matching second-order statistics between style and stylized features for style transfer, we conjecture that matching first-order mean vectors is also important, and this is supported by centralization and decentralization. We explain how it is supported in the Supplementary Materials using the algorithm of the prior transformations~\cite{huang2017arbitrary,li2017universal,lu2019closed}. 


Formally, let $\mathbf{\bar{F}}$ $=$ $\mathbf{F} - \mu(\mathbf{F})$ denote the centralized feature of $\mathbf{F}$,
where $\mu(\mathbf{F})$ is the mean vector across the $HW$ columns of $\mathbf{F}$ and the matrix-vector subtraction in $\mathbf{F}$ - $\mu(\mathbf{F})$ is done by array broadcasting. The algorithm applied by prior transformations~\cite{huang2017arbitrary,li2017universal,lu2019closed} is as follows: (1) Centralization: centralize $\mathbf{F}_c$ and $\mathbf{F}_s$ to be $\mathbf{\bar{F}}_c$ and $\mathbf{\bar{F}}_s$. (2) Second-order statistic matching: alter $\mathbf{\bar{F}}_c$ to be $\mathbf{\bar{F}}_t$ such that for AdaIN~\cite{huang2017arbitrary} the variances of $\mathbf{\bar{F}}_s$ and $\mathbf{\bar{F}}_t$ are equal, and for ZCA~\cite{li2017universal} and OST~\cite{lu2019closed} the covariances of $\mathbf{\bar{F}}_s$ and $\mathbf{\bar{F}}_t$ are equal. (3) Decentralization: add $\mu(\mathbf{F}_s)$ to $\mathbf{\bar{F}}_t$ to derive the transformed feature $\mathbf{F}_t$.


We modify IterFT to apply centralization before and decentralization after the iterative feature update. Mathematically, with the centralized content and style features $\mathbf{\bar{F}}_c$ and $\mathbf{\bar{F}}_s$, the modified optimization problem is written as Eq.~\ref{eq:mod_iterFT_obj}, where the constraint $\mu(\mathbf{\bar{F}}_t)=\vec{0}$ requires $\mathbf{\bar{F}}_t$ to be centralized. With $\mathbf{\bar{F}}_t$ initialized as $\mathbf{\bar{F}}_c$, the centralized feature $\mathbf{\bar{F}}_t$ can be solved using gradient descent\footnote{While Quasi-Newton methods seem plausible, they are impractical here since they are memory-intensive due to computing Hessian matrices; e.g., for FHD input (1920x1080), the Hessian matrix of the loss function in Eq.~\ref{eq:mod_iterFT_obj} has (64x1920x1080)$^2$ = 1.76e$^{16}$ elements at the $\textit{relu}1\_1$ layer.} in Eq.~\ref{eq:mod_iterFT_upd} with the analytical gradient $\frac{\mathrm{d}l}{\mathrm{d}\mathbf{\bar{F}}_t}$ in Eq.~\ref{eq:mod_iterFT_grad}. 
\begin{equation}
\small
    \begin{aligned}
    \min_{\mathbf{\bar{F}}_t} l(\mathbf{\bar{F}}_t) = &\min_{\mathbf{\bar{F}}_t} ||\mathbf{\bar{F}}_t - \mathbf{\bar{F}}_c||^2_2 + \lambda||\frac{1}{n_c}\mathbf{\bar{F}}_t\mathbf{\bar{F}}_t^\mathrm{T}-\frac{1}{n_s}\mathbf{\bar{F}}_s\mathbf{\bar{F}}_s^\mathrm{T}||^2_2 \\
    &\text{subject to } \mu(\mathbf{\bar{F}}_t) = 0,
    \label{eq:mod_iterFT_obj}
    \end{aligned}
\end{equation}
\begin{equation}
    \mathbf{\bar{F}}_t \leftarrow \mathbf{\bar{F}}_t - \eta \frac{\mathrm{d}l}{\mathrm{d}\mathbf{\bar{F}}_t},~\mathbf{\bar{F}}_t \leftarrow \mathbf{\bar{F}}_t - \mu(\mathbf{\bar{F}}_t),
    \label{eq:mod_iterFT_upd}
\end{equation}
\begin{equation}
    \frac{\mathrm{d}l}{\mathrm{d}\mathbf{\bar{F}}_t} = 2(\mathbf{\bar{F}}_t - \mathbf{\bar{F}}_c) + \frac{4\lambda}{n_c}(\frac{1}{n_c}\mathbf{\bar{F}}_t\mathbf{\bar{F}}_t^\mathrm{T}-\frac{1}{n_s}\mathbf{\bar{F}}_s\mathbf{\bar{F}}_s^\mathrm{T})\mathbf{\bar{F}}_t.
    \label{eq:mod_iterFT_grad}
\end{equation}
\noindent
The final resulting feature $\mathbf{F}_t$ is $\mathbf{\bar{F}}_t$ decentralized by $\mu(\mathbf{F}_s)$: $\mathbf{\bar{F}}_t$ $+$ $\mu(\mathbf{F}_s)$. We exemplify the improved quality resulting from our modification compared to IterFT in Fig.~\ref{fig:iter_vs_Moditer}.

Notice that the constraint that $\mu(\mathbf{\bar{F}}_t)=\vec{0}$ is always satisfied while gradient descent minimizes the loss value. This can be shown by first assuming the current $\mathbf{\bar{F}}_t$ is centralized. With this, we have $\mu(\mathbf{\bar{F}}_t - \eta \frac{\mathrm{d}l}{\mathrm{d}\mathbf{\bar{F}}_t})$ $=$ $\mu(\mathbf{\bar{F}}_t)-\eta\mu( \frac{\mathrm{d}l}{\mathrm{d}\mathbf{\bar{F}}_t})$ $=$ $-\eta\mu( \frac{\mathrm{d}l}{\mathrm{d}\mathbf{\bar{F}}_t})$. This implies the updated feature is centralized if $\mu(\frac{\mathrm{d}l}{\mathrm{d}\mathbf{\bar{F}}_t})$ $=\vec{0}$, which can be shown as follows: $\mu(\frac{\mathrm{d}l}{\mathrm{d}\mathbf{\bar{F}}_t})$ $=$ $2(\mu(\mathbf{\bar{F}}_t) - \mu(\mathbf{\bar{F}}_c))$ $+$ $\mu(\mathbf{S}\mathbf{\bar{F}}_t)$ $=$ $\mu(\mathbf{S}\mathbf{\bar{F}}_t)$ $=$ $\mathbf{S}\mu(\mathbf{\bar{F}}_t)$ $=\vec{0}$. Therefore, with $\mathbf{\bar{F}}_t$ initialized as $\mathbf{\bar{F}}_c$, which is centralized, the sequels of the updated $\mathbf{\bar{F}}_t$'s are all centralized.

\subsection{Line search-based feature transformation}
\label{sec:LS-FT}
An issue that remains in our Modified IterFT is that the algorithm requires multiple iterations to update the feature, which is slow. In practice, the values of the learning rate $\eta$ and the number of iterations $n_{upd}$ are empirically set to be $0.01$ and $15$, respectively, in~\cite{chiu2020iterative}. However, intuitively, the learning rate $\eta$ should be dynamically determined such that $\eta$ is larger in the beginning iterations to accelerate the convergence and smaller in the later to fine-tune the solution.\footnote{While a trivial solution to accelerate convergence for existing methods could be to simply increase the learning rate, this is insufficient since we can't know how much to increase the rate.  We discuss this in Supplementary Materials. } With a dynamic $\eta$, the number of iterations $n_{upd}$ can then be greatly reduced.  To dynamically determine the value of $\eta$ for a new iteration with the latest $\mathbf{\bar{F}}_t$ from the last iteration and the derivative $\frac{\mathrm{d}l}{\mathrm{d}\mathbf{\bar{F}}_t}$ calculated from Eq.~\ref{eq:mod_iterFT_grad}, we solve the following line search optimization problem:
\begin{equation}
    \min_{\eta} l(\mathbf{\bar{F}}_t - \eta\frac{\mathrm{d}l}{\mathrm{d}\mathbf{\bar{F}}_t})\text{ subject to }\eta>0,
    \label{eq:line_search}
\end{equation}
where the loss function $l$ is defined in Eq.~\ref{eq:mod_iterFT_obj}, and the constraint $\eta$ $>$ $0$ forces $\mathbf{\bar{F}}_t$ to move toward the descent direction. The meaning of Eq.~\ref{eq:line_search} is that we start from the point $\mathbf{\bar{F}}_t$ and search in the opposite direction of $\frac{\mathrm{d}l}{\mathrm{d}\mathbf{\bar{F}}_t}$, i.e. the descent direction, to find a new point $\mathbf{\bar{F}}_t - \eta\frac{\mathrm{d}l}{\mathrm{d}\mathbf{\bar{F}}_t}$ which minimizes the loss function. With the substitution of $\frac{\mathrm{d}l}{\mathrm{d}\mathbf{\bar{F}}_t}$ in Eq.~\ref{eq:line_search} with Eq.~\ref{eq:mod_iterFT_grad} and some arithmetic with calculus (detailed derivation is provided in Supplementary Materials), we can show that the optimal $\eta$ should be a solution to this cubic equation:

\begin{equation}
    a\eta^3 + b\eta^2 + c\eta + d = 0,
    \label{eq:line_search_cubic}
\end{equation}
with the coefficients defined as follows:
\begin{equation} 
\small
    a =  \frac{2\lambda}{n_c^2}\mathrm{tr}[\mathbf{D_2}\mathbf{D_2}],~b =  -\frac{6\lambda}{n_c^2}\mathrm{tr}[\mathbf{D}_\mathbf{F}\mathbf{D_2}],~d = -\frac{1}{2}\mathrm{tr}[\mathbf{D_2}],
    \label{eq:coefficient_a}
\end{equation}
\begin{equation} 
\small
    c = \mathrm{tr}[\mathbf{D_2}] +  \frac{2\lambda}{n_c}\mathrm{tr}[\mathbf{D_2}\mathbf{S}] + \frac{2\lambda}{n_c^2}\big(\mathrm{tr}[\mathbf{D}_\mathbf{F}\mathbf{D}_\mathbf{F}] + \mathrm{tr}[\mathbf{D}_\mathbf{F}\mathbf{D}_\mathbf{F}^\mathrm{T}] \big),
    \label{eq:coefficient_c}
\end{equation}
where $\mathbf{D_2} \equiv \mathbf{D}\mathbf{D}^\mathrm{T}$, $\mathbf{D}_\mathbf{F} \equiv \mathbf{D}\mathbf{\bar{F}}_t^\mathrm{T}$, $\mathbf{D}$ $\equiv$ $\frac{\mathrm{d}l}{\mathrm{d}\mathbf{\bar{F}}_t}$ and $\mathbf{S}$ $\equiv$ $\frac{1}{n_c}\mathbf{\bar{F}}_t\mathbf{\bar{F}}_t^\mathrm{T}-\frac{1}{n_s}\mathbf{\bar{F}}_s\mathbf{\bar{F}}_s^\mathrm{T}$.

Although Eq.~\ref{eq:coefficient_a} and Eq.~\ref{eq:coefficient_c} may look intimidating at the first glance, Eq.~\ref{eq:line_search_cubic} can actually be solved efficiently for the following reasons. First, $\mathbf{D}$ and its sub-term $\mathbf{S}$ already must be computed for the gradient descent calculation in Modified IterFT, and so do not introduce extra overhead when line-searching $\eta$. Second, since all coefficients are rooted in only two repeated terms $\mathbf{D_2}$ and $\mathbf{D}_\mathbf{F}$, we need to compute $\mathbf{D_2}$ and $\mathbf{D}_\mathbf{F}$ just once to derive all coefficients. Third, the matrix multiplications $\mathbf{D_2}$ and $\mathbf{D}_\mathbf{F}$ and the trace operation $\mathrm{tr}[\mathbf{A}\mathbf{B}]$ of two matrices $\mathbf{A}$ and $\mathbf{B}$ can be computed in parallel with a GPU. Finally, with the coefficients computed, we can solve the cubic function in Eq.~\ref{eq:line_search_cubic} in constant time using the cubic formula~\cite{cubic_equation}\footnote{Eq.~\ref{eq:line_search}
is a quartic function of $\eta$. If there are three positive solutions to Eq.~\ref{eq:line_search_cubic}, they correspond to a local minimum, a local maximum, and the global minimum of Eq.~\ref{eq:line_search}. Therefore, we just plug the solutions to Eq.~\ref{eq:line_search} to see which one results in the lowest value and that is the one we pick.}.

To comply with the constraint $\eta$ $>$ $0$, we have to ensure that there is at least one positive solution to Eq.~\ref{eq:line_search_cubic}. We prove this in Supplementary Materials.

In summary, our line search-based feature transformation (LS-FT) produces the transformed feature $\mathbf{F}_t$ with the content and style features $\mathbf{F}_c$ and $\mathbf{F}_s$ in four steps: 
\begin{enumerate}
\setlength\itemsep{0.05em}
\item Centralize the content and style features to  be $\mathbf{\bar{F}}_c$ and $\mathbf{\bar{F}}_s$ and initialize $\mathbf{\bar{F}}_t$ to be $\mathbf{\bar{F}}_c$. 
\item Calculate the gradient from Eq.~\ref{eq:mod_iterFT_grad} and the learning rate $\eta$ from Eq.~\ref{eq:line_search_cubic}. 
\item Update $\mathbf{\bar{F}}_t$ according to Eq.~\ref{eq:mod_iterFT_upd} and iterate from the step 2 if needed. 
\item Decentralize $\mathbf{\bar{F}}_t$ by adding the mean $\mu(\mathbf{F}_s)$ of the style feature. 
\end{enumerate}
We will show in Sec.~\ref{sec:exp_convergence} that, unlike Modified IterFT, iteration in step 3 is not necessary and one feature update is sufficient for LS-FT. To balance the magnitude of the content loss $||\mathbf{\bar{F}}_t - \mathbf{\bar{F}}_c||^2_2$ and style loss $\lambda||\frac{1}{n_c}\mathbf{\bar{F}}_t\mathbf{\bar{F}}_t^\mathrm{T}-\frac{1}{n_s}\mathbf{\bar{F}}_s\mathbf{\bar{F}}_s^\mathrm{T}||^2_2$ in Eq.~\ref{eq:mod_iterFT_obj}, it is best to have the value of $\lambda$ close to the ratio $\frac{||\mathbf{\bar{F}}_t - \mathbf{\bar{F}}_c||^2_2}{||\frac{1}{n_c}\mathbf{\bar{F}}_t\mathbf{\bar{F}}_t^\mathrm{T}-\frac{1}{n_s}\mathbf{\bar{F}}_s\mathbf{\bar{F}}_s^\mathrm{T}||^2_2}$. However, since $\mathbf{\bar{F}}_t$ is unknown, we replace it with $\mathbf{0}$ and set $\lambda$ to be $\frac{||\mathbf{\bar{F}}_c||^2_2}{||\frac{1}{n_s}\mathbf{\bar{F}}_s\mathbf{\bar{F}}_s^\mathrm{T}||^2_2}$ for content-style balance. 

For tuning the content style balance (i.e., determined by the value of $\lambda$), we introduce a coefficient $\alpha$ such that $\lambda$ $=$ $\alpha\frac{||\mathbf{\bar{F}}_c||^2_2}{||\frac{1}{n_s}\mathbf{\bar{F}}_s\mathbf{\bar{F}}_s^\mathrm{T}||^2_2}$.  Varying $\alpha$ in turn supports boosting the stylization strength or content preservation. We show results in the main paper for fixed $\alpha$'s and expanded analysis of the effect of $\alpha$ in the Supplementary Materials.

\section{Experiments}

We now evaluate our transformation's ability to generalize in establishing a style-content balance  across multiple photorealistic style transfer architectures (Sec.~\ref{sec:exp_content_style_balance}) and its improved computational efficiency (Sec.~\ref{sec:exp_speed}).  

\subsection{Convergence of line search optimization}
\label{sec:exp_convergence}
We first describe our analysis to establish how many iterations to use for our line search optimization in our LS-FT transformation, by embedding it in four style transfer architectures: WCT$^2$, PhotoWCT, PhotoWCT$^2$, and PCA-d. 

\vspace{-3.5mm}
\paragraph{Implementations.} 
The baseline is Modified IterFT, which we accelerate by introducing line search optimization to it. Modified IterFT follows IterFT~\cite{chiu2020iterative} to adopt $0.01$ as the learning rate and $15$ as the number of iterations for each layer. We want to know how many iterations are required for LS-FT to outperform the performance of Modified IterFT after 15 iterations. 

\vspace{-3.5mm}
\paragraph{Dataset.} 
We use the PST dataset~\cite{xia2020joint}, which is the largest dataset for photorealistic style transfer evaluation.  It consists of 786 pairs of a content image and a style image.

\vspace{-3.5mm}
\paragraph{Metric.}
Recall that both Modified IterFT and LS-FT iteratively adapt the content feature with respect to the style feature at each $\textit{reluN\_1}$ layer for 15 iterations to minimize the loss value defined in Eq.~\ref{eq:mod_iterFT_obj}. We monitor the loss value after each iteration of a transformation. Specifically, for each transformation there is a series of 15 loss values calculated for each $\textit{reluN\_1}$ layer. Taking all input pairs into account, we have 786 series of 15 loss values at each $\textit{reluN\_1}$ layer. We compute the mean and standard deviation across the 786 series and plot the mean series as a curve and the standard deviation series as a shaded area around the curve.

\begin{figure*}[!t]
    \centering
    \includegraphics[width=0.9\textwidth]{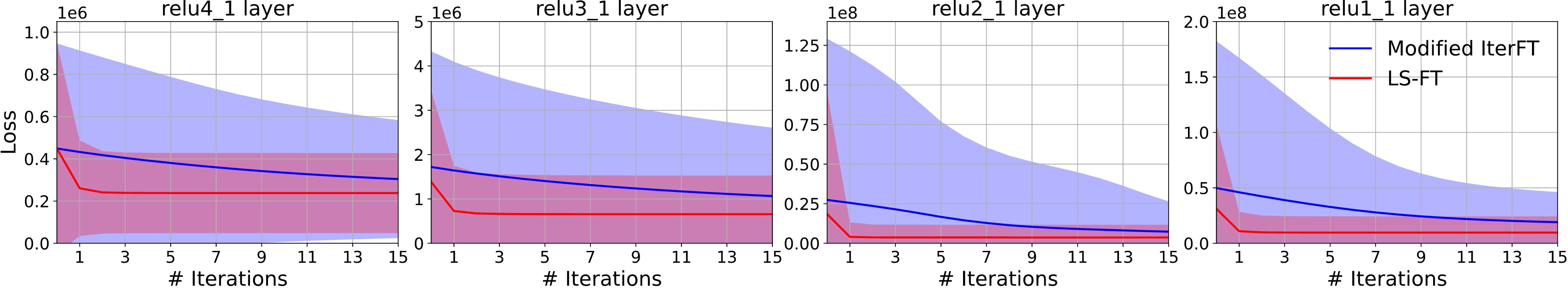}
    \vspace{-0.25em}
    \caption{\small Convergence comparison between Modified IterFT and LS-FT. Here Modified IterFT and LS-FT are tested on the PhotoWCT$^2$ model~\cite{chiu2021photowct2} which applies feature transformations at the bottleneck ($\textit{relu4\_1}$ layer) and the $\textit{relu3\_1}$, $\textit{relu2\_1}$, $\textit{relu1\_1}$ layers in the decoder. For a content-style input pair, a loss value is calculated according to Eq.~\ref{eq:mod_iterFT_obj} after each iteration of Modified IterFT or LS-FT at each layer, resulting in a series of 15 loss values for 15 iterations. A curve shows the mean series across 786 series from all 786 input pairs. The surrounding shaded area indicates the region within one standard deviation. It is observed that LS-FT converges faster than Modified IterFT and only one iteration is sufficient for LS-FT to outperform Modified IterFT at each layer.}
    \label{fig:convergence_rate}
\end{figure*}

\vspace{-3.5mm}
\paragraph{Results.} 
Mean loss curves and associated standard deviations are shown in Fig.~\ref{fig:convergence_rate} for each of the four $\textit{reluN\_1}$  layers for PhotoWCT$^2$. Due to limited space and similar findings, results for WCT$^2$, PhotoWCT, and PCA-d are provided in the Supplementary Materials. 

When comparing LS-FT to Modified IterFT, we observe LS-FT converges much faster.  For example, observing the mean curves at the $\textit{relu4\_1}$ layer, LS-FT converges in only two iterations with the second iteration slightly improving from the first while Modified IterFT slowly converges over all iterations.  Moreover, the loss value of LS-FT after the first iteration is already lower than that of Modified IterFT after 15 iterations, suggesting that one iteration is sufficient for LS-FT at the $\textit{relu4\_1}$ layer. Note that the smaller standard deviation of LS-FT than that of Modified IterFT at each iteration implies the faster convergence of LS-FT is universal across the dataset. We notice the same phenomenon occurs at the other three layers, suggesting one feature update is enough for LS-FT at these layers as well.

\subsection{Content preservation and stylization strength}

\label{sec:exp_content_style_balance}
To evaluate our feature transformation's ability to achieve a better balance between content preservation and stylization strength than prior transformations, we benchmark multiple transformations embedded in four autoencoder-based style transfer models: WCT$^2$~\cite{yoo2019photorealistic}, PhotoWCT~\cite{li2018closed}, PhotoWCT$^2$~\cite{chiu2021photowct2}, and PCA-d~\cite{chiu2022pca}. 

\vspace{-3.5mm}
\paragraph{Dataset.} 
We again test with the PST dataset~\cite{xia2020joint}, which consists of 786 pairs of a content image and a style image.

\vspace{-3.5mm}
\paragraph{Our Implementations.} 
We again evaluate our LS-FT and Modified IterFT (i.e., our stable IterFT~\cite{chiu2020iterative}).

\vspace{-3.5mm}
\paragraph{Baselines.} 
For comparison, we evaluate ZCA~\cite{li2017universal}, OST~\cite{lu2019closed}, AdaIN~\cite{huang2017arbitrary}, and MAST~\cite{huo2021manifold}.  IterFT~\cite{chiu2020iterative} is not used for comparison since it results in many unreasonable results, as is shown in the Supplementary Materials.

\vspace{-3.5mm}
\paragraph{Metrics.} 
For each model-transformation pair, we compute the mean content loss and mean style loss across all stylized images by computing for each stylized image $I_{sty}$, its content loss from the content image $I_c$ and style loss from the style image $I_s$. To define the losses, we let $\mathbf{\bar{F}}_{k,N}$ $\in$ $C_N$ $\times$ $H_{k,N} W_{k,N}$  ($N$ $=$ $1,2,3,4$) denote the centralized $\textit{reluN\_1}$ feature of the image $I_k$ ($k$ $\in$ $\{c,s,sty\}$). Following NST~\cite{gatys2016image}, which uses the $\textit{relu4\_1}$ layer to represent content and multiple layers (e.g. $\textit{reluN\_1}$, N $=$ $1,2,3,4$) to represent style, we define the content loss as $||\mathbf{\bar{F}}_{sty,4}-\mathbf{\bar{F}}_{c,4}||_2^2$, and the style loss as $\sum_{N=1}^4||\frac{1}{H_{c,N}W_{c,N}}\mathbf{\bar{F}}_{sty,N}\mathbf{\bar{F}}_{sty,N}^\mathrm{T}-\frac{1}{H_{s,N}W_{s,N}}\mathbf{\bar{F}}_{s,N}\mathbf{\bar{F}}_{s,N}^\mathrm{T}||_2^2$. 

Then, we evaluate the quality of stylized images using image quality assessment metrics: SSIM~\cite{wang2004image}, FSIM~\cite{zhang2011fsim}, and NIMA~\cite{talebi2018nima}. SSIM and FSIM assess the structural similarity between a stylized image and its source content image, while NIMA evaluates the stylized image as a standalone photo without referencing to the content image.

\begin{figure*}[!t]
    \centering
    \includegraphics[width=\linewidth]{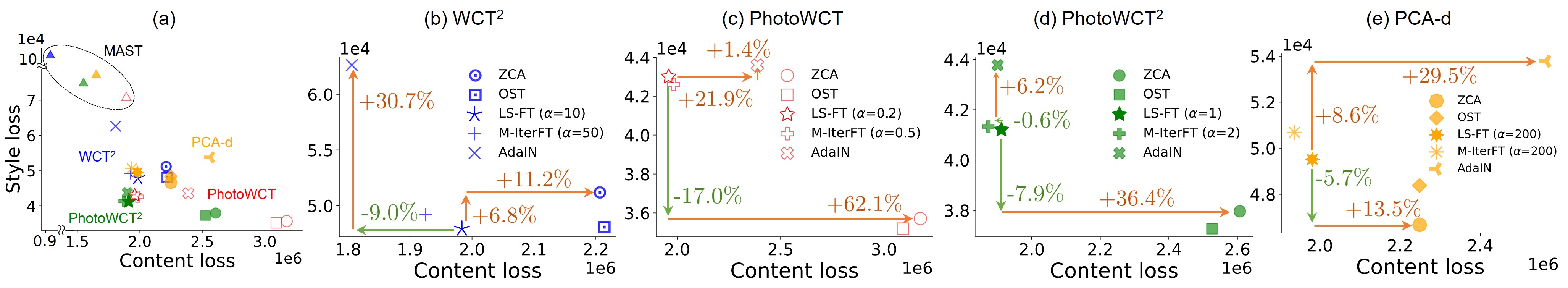}
    \vspace{-2em}
    \caption{Mean content losses vs. mean style losses of stylized images for different model-transformation pairs. Unlike other transformations, our Modified IterFT (M-IterFT) and LS-FT boosts WCT$^2$'s stylization strength, enhances PhotoWCT and PCA-d's content preservation, and reaches a good balance between style strength and content preservation for PhotoWCT$^2$.}
    \label{fig:content_style_losses}
\end{figure*}

\begin{figure*}[!t]
    \centering
    \includegraphics[width=0.8\textwidth]{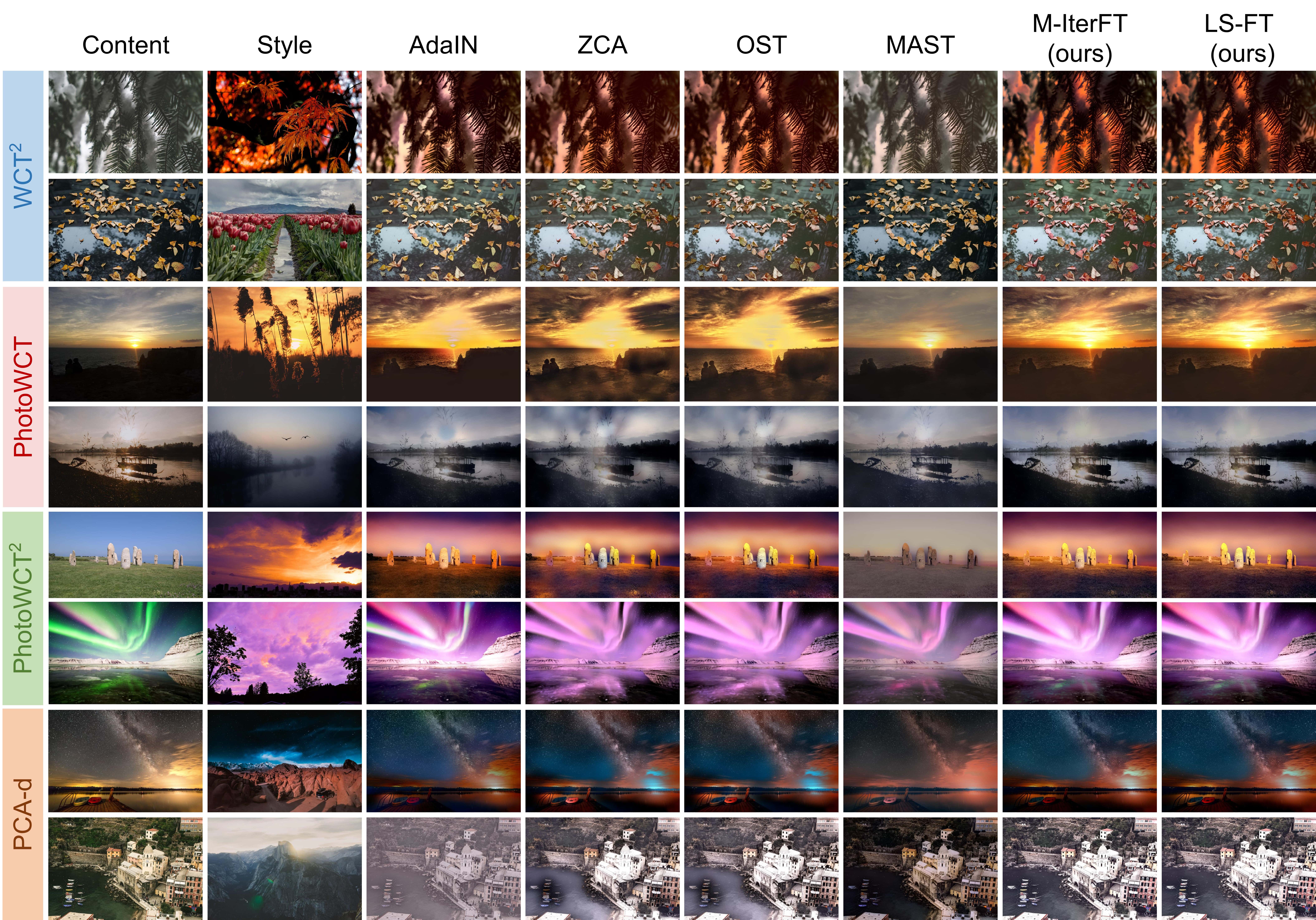}
    \vspace{-0.5em}
    \caption{Examples of a better content-style control from our transformations: Modified IterFT (M-IterFT) and LS-FT. Unlike other transformations, Modified IterFT and LS-FT can generalize to different models to achieve content-style balance.}
    \label{fig:qualitative_results}
\end{figure*}

\vspace{-3.5mm}
\paragraph{Results.} 
Fig.~\ref{fig:content_style_losses}(a) shows the overall distribution of mean content-style losses resulting from different model-transformation pairs. We observe that MAST results in the weakest style effects across all models, which is reinforced by qualitative results shown in Fig.~\ref{fig:qualitative_results}. Consequently, for our fine-grained analysis with respect to each model (Fig.~\ref{fig:content_style_losses}(b,c,d)), we exclude MAST from the analysis. 

As shown in Fig.~\ref{fig:content_style_losses}(b,c,d), our Modified IterFT and LS-FT have similar performance and consistently lead to a better balance between content preservation and stylization strength than prior transformations for all four style transfer architectures.  For WCT$^2$, its low content losses in Fig.~\ref{fig:content_style_losses}(a) come from its model design while it relies on a transformation to boost its stylization strength to faithfully reflect the style. We observe that LS-FT with $\alpha$ $=$ $10$ successfully boosts the stylization strength of WCT$^2$ compared to other transformations (Fig.~\ref{fig:content_style_losses}(b)), with AdaIN resulting in a $30.7\%$ greater style loss, and ZCA resulting in a $6.8\%$ greater style loss. Their worse style losses are also reflected in the qualitative results. As exemplified in the first row of Fig.~\ref{fig:qualitative_results}, while AdaIN poorly transfers the red leaf effect and ZCA transfers it more strongly, LS-FT improves upon ZCA by rendering more red color to the leaves.  

The PhotoWCT model is designed to reflect the style well and so have lower style losses, as demonstrated in Fig.~\ref{fig:content_style_losses}(a).  Consequently, it needs a transformation to boost its content preservation. Quantitatively, our transformations (e.g., LS-FT with $\alpha$ $=$ $0.2$) result in the lowest content loss compared to other transformations (Fig.~\ref{fig:content_style_losses}(c)), with ZCA resulting in a $62.1\%$ greater content loss, and AdaIN resulting in a $21.9\%$ greater content loss. Qualitatively, if we take the first row of the PhotoWCT panel in Fig.~\ref{fig:qualitative_results} for example, $62.1\%$ greater content loss of ZCA results in severe artifacts (uneven reflection of sunlight on the ground), while $21.9\%$ greater content loss of AdaIN makes it fail to preserve the finer content (less realistic sunset in the background).  

The PhotoWCT$^2$ model is designed to improve the content preservation from PhotoWCT and the stylization strength from WCT$^2$. However, it does not always preserve content well and transfer enough style effects, as mentioned before in Fig.~\ref{fig:banner} where it transfers insufficient style effects with AdaIN and introduces artifacts that ruin the content with ZCA. Thus, it needs a transformation to result in stronger stylization strength than AdaIN and better content preservation than ZCA. We observe our LS-FT with $\alpha$ $=$ $1$ can achieve this requirement by reducing the style loss of AdaIN by $6.2\%$ and the content loss of ZCA by $36.4\%$ (Fig.~\ref{fig:content_style_losses}(d)). As exemplified in the first row of the PhotoWCT$^2$ panel in Fig.~\ref{fig:qualitative_results}, our LF-FT preserves better content than ZCA and transfers stronger style than AdaIN.

PCA-d is an improved version of PhotoWCT$^2$ in that it is lightweight and achieves better content-style balance than PhotoWCT$^2$. While \cite{chiu2022pca} shows that PCA-d produces realistic results and reflects good style effects when ZCA is used as the transformation, our experiment shows that ZCA can still produce slight artifacts (unnatural sunlight in the first row of the PCA-d panel in Fig.~\ref{fig:qualitative_results}) and severe artifacts occasionally (misty artifacts on the lake in the second row). As shown in Fig.~\ref{fig:content_style_losses}(e), AdaIN performs even worse than ZCA with worse content and style losses. Our LS-FT, in contrast, mitigates the artifacts by reducing the content loss of ZCA by $13.5\%$, as exemplified in Fig.~\ref{fig:qualitative_results}.



\begin{figure*}[!t]
    \centering
    \includegraphics[width=\textwidth]{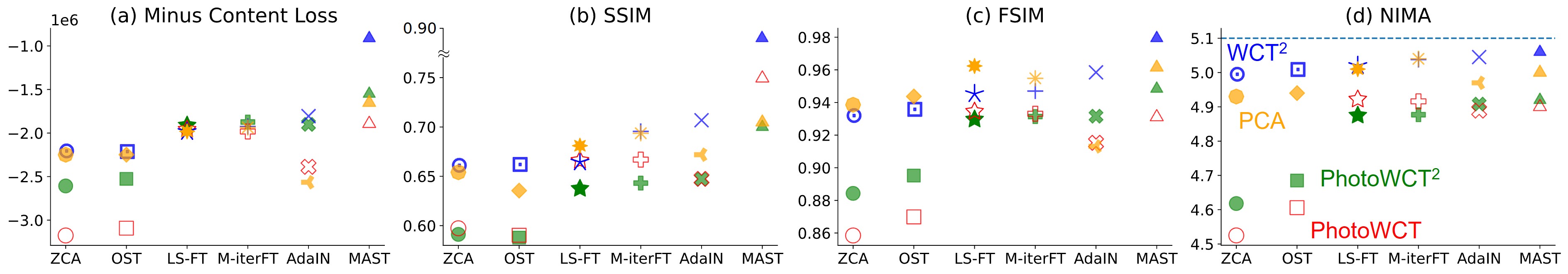}
    \vspace{-1.75em}
    \caption{\small Quality assessment of stylized images resulting from different model-transformation pairs. For each model, compared to the popular AdaIN and ZCA, our Modified IterFT and LS-FT can result in a score that is comparable to that of AdaIN and higher than that of ZCA. Note that the dash line in (d) indicates the average NIMA score for content imges in the PST dataset.}
    \label{fig:image_quality}
\end{figure*}

The mean quality scores of stylized images resulting from different model-transformation pairs reinforce our aforementioned findings (Fig.~\ref{fig:image_quality}).  In particular, for PhotoWCT and PCA-d the quality scores of LS-FT and Modified IterFT are the highest of all transformations except MAST due to the boost of content preservation, while in the cases of the other two models, LS-FT, Modified IterFT, and AdaIN have comparable quality scores which are higher than those from ZCA and OST. Note that MAST tends to have higher scores since it usually weakly adapts content.

\subsection{Speed performance}
\label{sec:exp_speed}
We finally compare the speed of our LS-FT to other transformations, when embedded in four style transfer architectures: WCT$^2$~\cite{yoo2019photorealistic}, PhotoWCT~\cite{li2018closed}, PhotoWCT$^2$~\cite{chiu2021photowct2}, and PCA-d~\cite{chiu2022pca}.  All tests are conducted with an Nvidia-1080Ti GPU with 11GB memory.

\vspace{-3.5mm}
\paragraph{Our Implementations.} 
We evaluate our LS-FT transformation and its ablated variant that lacks line search for the speed up: Modified IterFT (stable IterFT~\cite{chiu2020iterative}).

\vspace{-3.5mm}
\paragraph{Baselines.} 
We evaluate ZCA~\cite{li2017universal}, OST~\cite{lu2019closed},  AdaIN~\cite{huang2017arbitrary}, and MAST~\cite{huo2021manifold}.\footnote{IterFT has almost the same speed of M-IterFT and so is ignored here.} 

\vspace{-3.5mm}
\paragraph{Dataset.}
We test on four resolutions: $1280$ $\times$ $720$ (HD), $1920$ $\times$ $1080$ (FHD), $2560$ $\times$ $1440$ (QHD), and $3840$ $\times$ $2160$ (UHD or 4K). To collect images, we downloaded a 4K video~\cite{4kvideo} from YouTube, sampled 100 frames, and downsampled each frame to the other lower resolutions. For each model, the speed of a transformation for each resolution is averaged across the total 100 images. 

\vspace{-2mm}
\paragraph{Results.}
Tab.~\ref{tab:speed} shows the stylization speeds. We report results from PCA-d in another table in the Supplementary Materials because PCA-d is a distilled model that produces lightweight features leading to faster transformations and because of limited space.

\begin{table}[!t]
\setlength{\tabcolsep}{2.0pt}
    \centering
    \fontsize{8}{10}\selectfont
    \begin{tabular}{c c c c c c c c}
    \toprule
    {} & \multirow{2}{*}{\makecell{WCT$^2$ / \\PhotoWCT / \\PhotoWCT$^2$}} & \multicolumn{4}{c}{Not Tunable} & \multicolumn{2}{c}{Tunable} \\
    \cmidrule(lr){3-6}\cmidrule(lr){7-8}
     &  & ZCA & OST & AdaIN & MAST & M-IterFT & LS-FT \\ [2mm]\midrule
    HD & 0.18 / 0.37 / 0.13 & 0.18 & 0.27 & \bf{0.01} & 0.60 & 0.29 & \bf{0.04} \\
    FHD & 0.59 / 0.63 / 0.24 & 0.20 & 0.29 & \bf{0.02} & 1.14 & 0.73 & \bf{0.09} \\
    QHD & OOM / 0.98 / 0.40 & 0.24 & 0.33 & \bf{0.06} & 2.15 & 1.27 & \bf{0.17} \\
    UHD & OOM / OOM / 0.88 & 0.33 & 0.40 & \bf{0.13} & 5.02 & 2.84 & \bf{0.34} \\
    \bottomrule
    \end{tabular}
    \vspace{-0.25em}
    \caption{\small The speeds for stylization of images of different resolutions using different transformations. For clarity, the time spent on the model and the transformation is separated. The transformations are categorized into two groups based on whether they consider model adpativeness or not. LS-FT is consistently 7-8x times faster than Modified IterFT (M-IterFT) in all resolutions due to no need of multiple iterations. LS-FT also runs faster than or comparably to ZCA and OST. OOM: Out of Memory. Unit: Second.}
    \label{tab:speed}
\end{table}

Compared to its ablated variant Modified IterFT which also controls the balance between stylization strength and content preservation, LS-FT is consistently 7-8x times faster due to no need of multiple iterations. For example, it takes LS-FT 0.34 seconds to stylize a UHD image, which is 8.35x faster than the 2.84 seconds for Modified IterFT. 

When comparing LS-FT to the four baseline transformations lacking control over the balance between stylization strength and content preservation, overall LS-FT is competitive.  For example, LS-FT is faster than OST and MAST in all resolutions, while LS-FT is comparably fast to ZCA in the UHD resolution case and faster than ZCA in the others. The one exception is AdaIN, which is the fastest transformation.  This is due to its simplest math formulation.

\section{Conclusion}
We derived a new line search-based feature transformation (LS-FT) for photorealistic style transfer. Experiments show LS-FT with different style transfer architectures outperforms existing transformations by either boosting stylization strength, enhancing photorealism, or reaching a better style-content balance, while running at a fast speed. 


\section*{Supplementary Materials}
This document supplements the main paper with the following.
\begin{enumerate}
    \item Experiments showing insufficiency of LST and DSTN for photorealistic style transfer (supplements Section 2 of the main paper).
    \item Insufficiency of linear interpolation for content-style control (supplements Section 3 of the main paper).
    \item Experiments which show that removing centralization and decentralization from AdaIN and ZCA leads to worse image quality (supplements Section 3.2 of the main paper).
    \item Explanation for how centralization and decentralization support mean vector matching (supplements Section 3.2 of the main paper).
    \item Derivation of Equation 7 in the main paper.
    \item Proof of at least one positive solution to Equation 7 in the main paper.
    \item Computation of the values of $\eta$ searched by LS-FT (supplements Section 3.3 of the main paper).
    \item Qualitative results demonstrating the effect of the content-style control knob $\alpha$ (supplements Section 3.3 of the main paper).
    \item Convergence comparison between Modified IterFT and LS-FT on WCT$^2$, PhotoWCT, and PCA-d (supplements Section 4.1 of the main paper).
    \item Qualitative results showing that our approaches fix the unreasonable results from IterFT (supplements Section 4.2 of the main paper).
    \item Speed of transformations on PCA-d (supplements Section 4.3 of the main paper).
\end{enumerate}

\section*{Insufficiency of LST for photorealistic style transfer}

The prior work of LST~\cite{li2019learning} and DSTN~\cite{hong2021domain} claim their autoencoder-based models can be used for photorealistic style transfer. However, they did not provide strong quantitative analysis to support the claim. Here we show that they are insufficient for photorealistic style transfer with the quantitative and qualitative evidence showing they preserve content poorly.

\begin{figure*}[!h]
    \centering
    \includegraphics[width=\linewidth]{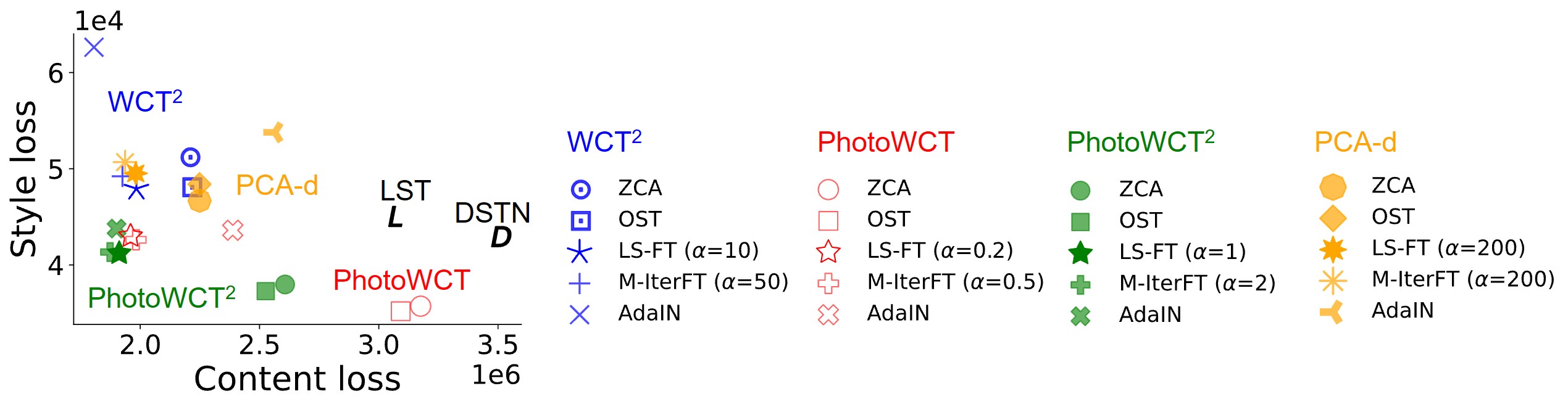}
    \caption{LST and DSTN do not preserve content and photorealism well compared to most transformation-model pairs.}
    \label{fig:lst_performance}
\end{figure*}

\begin{figure*}[t]
    \centering
    \includegraphics[width=\textwidth]{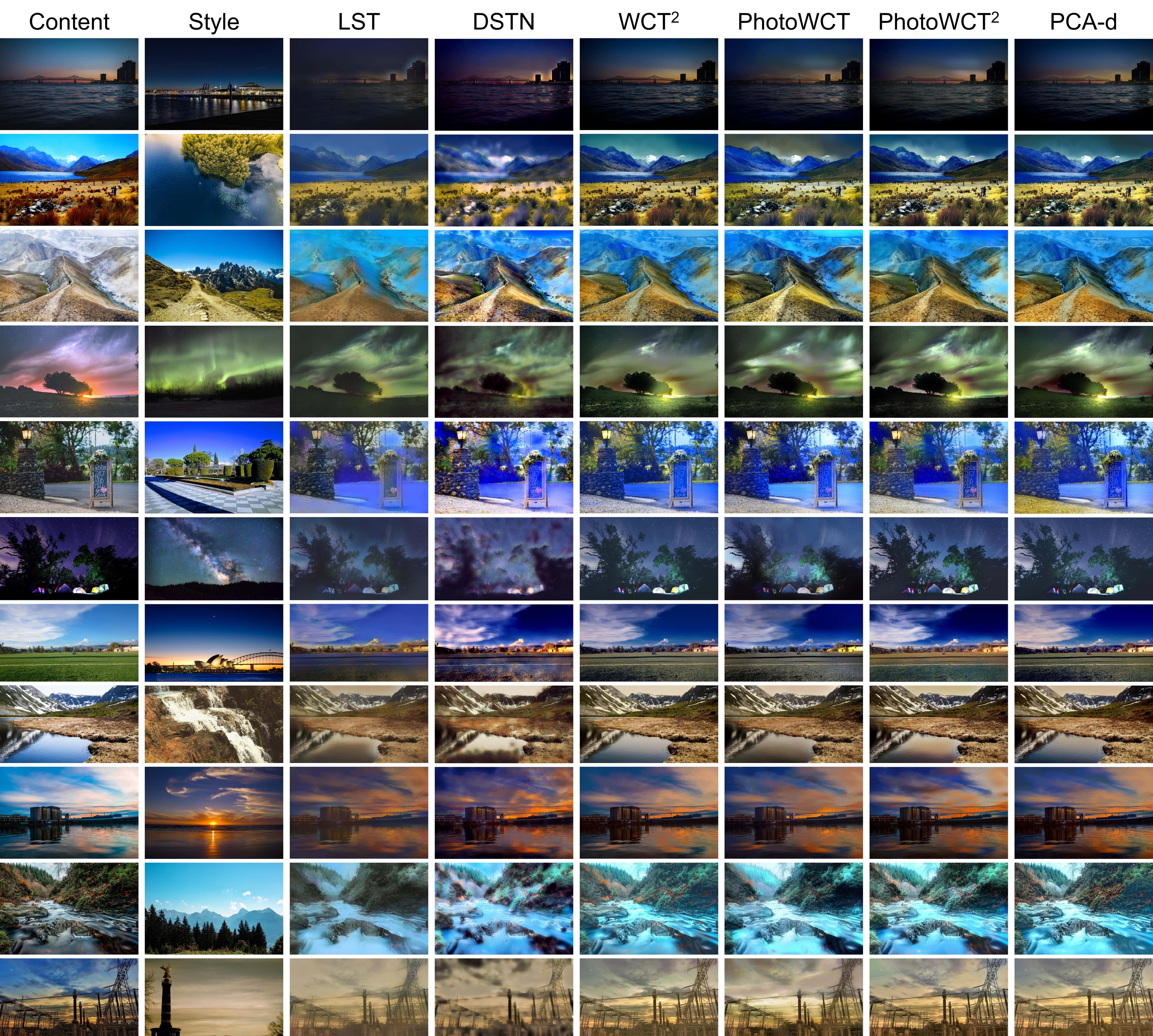}
    \caption{Qualitative comparison of LST~\cite{li2019learning} and DSTN~\cite{hong2021domain} to WCT$^2$~\cite{yoo2019photorealistic}, PhotoWCT~\cite{li2018closed}, PhotoWCT$^2$~\cite{chiu2021photowct2}, and PCA-d~\cite{chiu2022pca} with our LS-FT as the feature transformation. The results show that LST and DSTN produce poor photorealism.}
    \label{fig:comparison_to_LST}
\end{figure*}

First, we notice from Fig.~\ref{fig:lst_performance} that LST results in almost as bad content preservation as PhotoWCT with ZCA as the feature transformation and DSTN has an even worse content loss. Qualitatively, as exemplified in Fig.~\ref{fig:comparison_to_LST}, compared to the results from WCT$^2$~\cite{yoo2019photorealistic}, PhotoWCT~\cite{li2018closed},  PhotoWCT$^2$~\cite{chiu2021photowct2}, and PCA-d~\cite{chiu2022pca} with our LS-FT as the feature transformation, the results from LST are prone to blurred boundaries (low sharpness) and dullness (low contrast) and the results from DSTN have many severe artifacts.

\clearpage

\clearpage

\begin{figure}[!t]
    \centering
    \includegraphics[width=\linewidth]{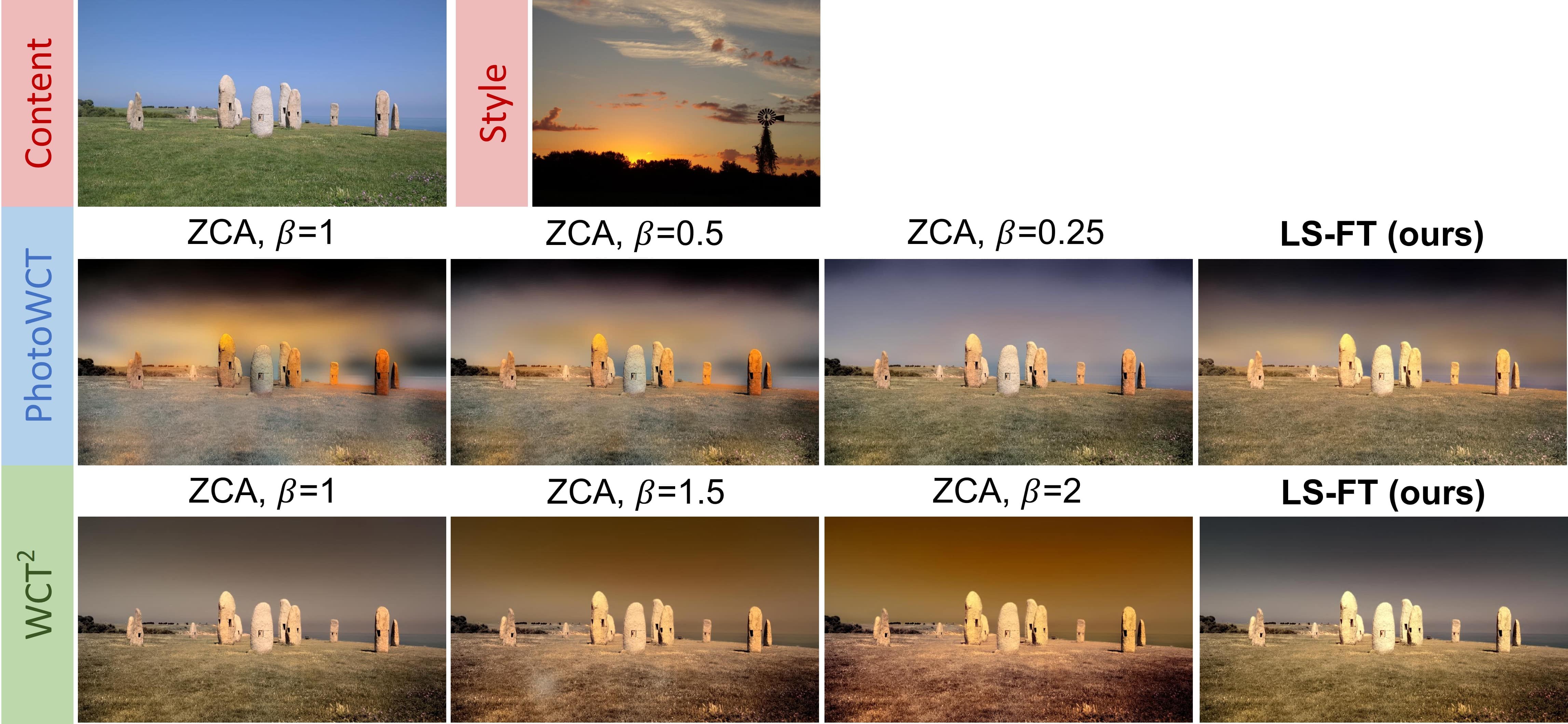}
    \caption{Insufficiency of linear interpolation for content-style control.}
    \label{fig:linear_interpolation}
\end{figure}

\section*{Insufficiency of linear interpolation for content-style control}
In \cite{li2017universal}, where ZCA is proposed for style transfer, the authors also propose to use linear interpolation between the style feature $\mathbf{F}_s$ and the transformed feature $\mathbf{F}_{t}$ to realize content-style controllability. That is, the feature to be decoded is $\beta\mathbf{F}_{t}+(1-\beta)\mathbf{F}_s$, where $\beta$ is the parameter that controls style strength. Mathematically, such a linear interpolation is insufficient since the content-style relation is a non-linear function as described by Eq.~(1) in our main paper. Qualitatively, Fig.~\ref{fig:linear_interpolation} shows that when trying to reduce reduce artifacts that come from strong stylization strength for PhotoWCT by reducing $\beta$, both style effects and artifacts disappear together.  Alternatively, when trying to strengthen stylization strength for WCT$^2$ by increasing $\beta$, we can observe distorted style effects (Fig.~\ref{fig:linear_interpolation}). In contrast, our LS-FT can maintain the style effects well while controlling the balance between content and style.

\section*{Removing centralization and decentralization from AdaIN and ZCA leads to worse image quality}
In Section 3.2 of the main paper, we add centralization and decentralization to stabilize the performance of IterFT~\cite{chiu2020iterative}. Here we conduct an ablation study of removing centralization and decentralization from AdaIN~\cite{huang2017arbitrary} and ZCA~\cite{li2017universal} and test the resulting performance for the PhotoWCT$^2$ model~\cite{chiu2021photowct2}. The results are shown in Fig.~\ref{fig:ablated_adain_zca}. These reveal that the ablated AdaIN and the ablated ZCA may suffer from incomplete stylization, where parts of a content image receive limited to no stylization. Consequently, bad results may occur. This offers promising evidence that centralization and decentralization play an important role in synthesizing reasonable images.

\section*{Explanation for how centralization and decentralization support mean vector matching}
While prior works~\cite{gatys2016image,li2017demystifying} focus on explaining the reason of matching second-order statistics between style and stylized features for style transfer, we conjecture that matching first-order mean vectors is also important, and this is supported by centralization and decentralization. We illustrate this here for the ZCA algorithm and the same argument applies to OST and AdaIN.

Let $\mathbf{F}_{c/s}$ be the content/style feature and $\mu_{c/s}$ and $\mathbf{C}_{c/s}$$=$$\mathbf{\bar{F}}_{c/s}\mathbf{\bar{F}}_{c/s}^\mathrm{T}$ be their mean vectors and covariance matrices. The ZCA transformed feature $\mathbf{F}_{t}$ is given by $\mathbf{C}_s^{\frac{1}{2}}$$\mathbf{C}_c^{\frac{-1}{2}}$$\mathbf{\bar{F}}_{c}$$+$$\mu_s$. It can be shown that the mean and the covariance of $\mathbf{F}_t$ are exactly those of $\mathbf{F}_s$. However, without centralization and decentralization and replacing $\mathbf{C}_{c/s}$  by the gram matrix $\mathbf{G}_{c/s}$$=$$\mathbf{F}_{c/s}\mathbf{F}_{c/s}^\mathrm{T}$, the transformed feature becomes $\mathbf{F}_g$$=$$\mathbf{G}_s^{\frac{1}{2}}$$\mathbf{G}_c^{\frac{-1}{2}}$$\mathbf{F}_{c}$. Now only the gram matrices of $\mathbf{F}_g$ and $\mathbf{F}_s$ match, while their mean vectors differ. This may explain the results in Fig.~\ref{fig:ablated_adain_zca} here in the Supplementary Materials. An interesting area for future work is to establish \emph{why} mean matching is important.

\clearpage
\begin{figure*}[t]
    \centering
    \includegraphics[width=0.9\textwidth]{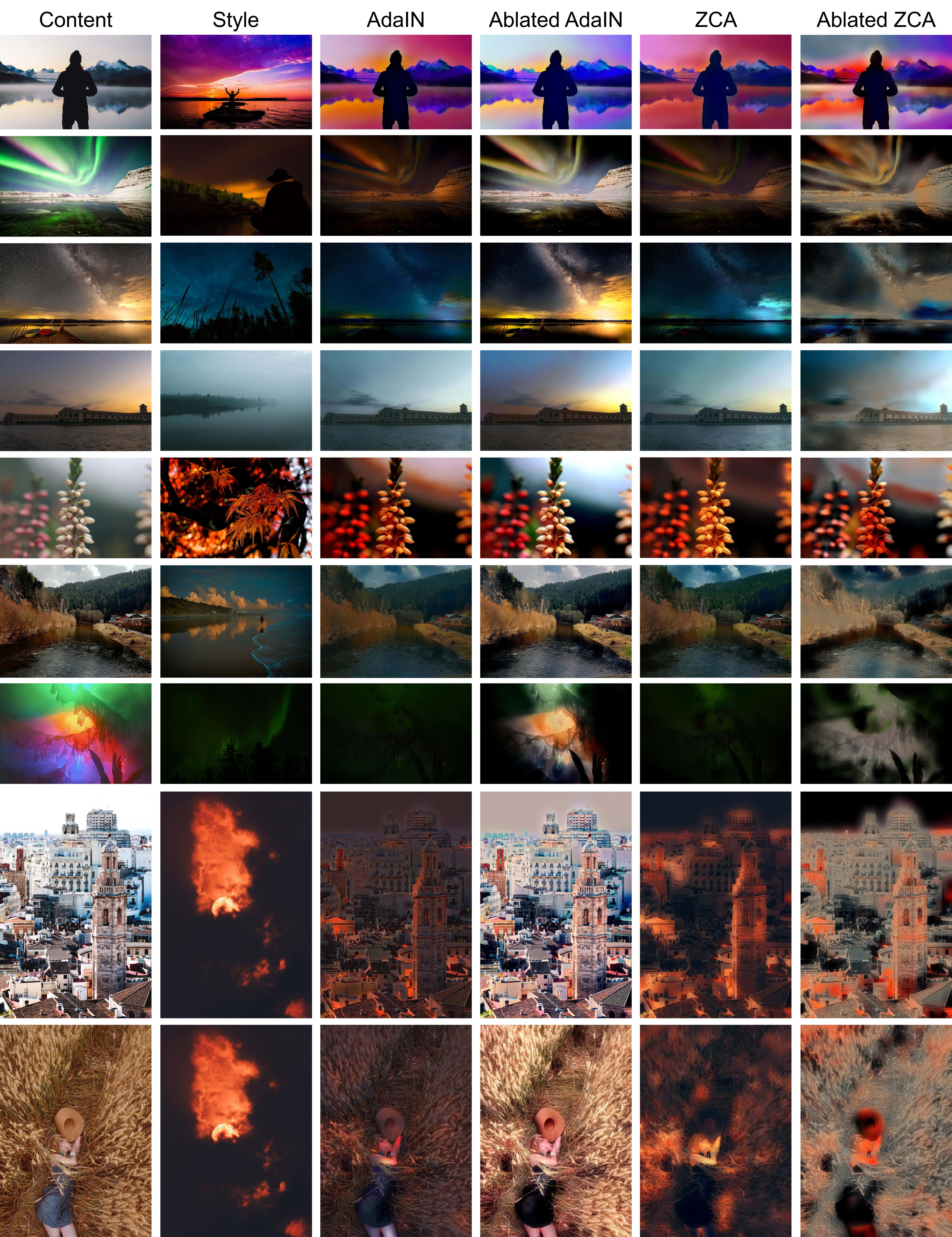}
    \caption{Ablation study of removing centralization and decentralization from AdaIN~\cite{huang2017arbitrary} and ZCA~\cite{li2017universal}. The results shows that centralization and decentralization play an important role in synthesizing reasonable images. The study is done with the PhotoWCT$^2$ model~\cite{chiu2021photowct2}.}
    \label{fig:ablated_adain_zca}
\end{figure*}
\clearpage

\section*{Derivation of Equation 7 in the main paper}
We show the following optimization problem:
\begin{equation}
    \min_{\eta} l(\mathbf{\bar{F}}_t - \eta\frac{\mathrm{d}l}{\mathrm{d}\mathbf{\bar{F}}_t}),
    \label{eq:optimization_problem}
\end{equation}
where:
\begin{equation}
    l(\mathbf{\bar{F}}_t) = \underbrace{||\mathbf{\bar{F}}_t - \mathbf{\bar{F}}_c||^2_2}_{\text{content loss}} + \lambda\underbrace{||\frac{1}{n_c}\mathbf{\bar{F}}_t\mathbf{\bar{F}}_t^\mathrm{T}-\frac{1}{n_s}\mathbf{\bar{F}}_s\mathbf{\bar{F}}_s^\mathrm{T}||^2_2}_{\text{style loss}}
    \label{eq:loss_function}
\end{equation}
and: 
\begin{equation}
    \frac{\mathrm{d}l}{\mathrm{d}\mathbf{\bar{F}}_t} = 2(\mathbf{\bar{F}}_t - \mathbf{\bar{F}}_c) + \frac{4\lambda}{n_c}(\frac{1}{n_c}\mathbf{\bar{F}}_t\mathbf{\bar{F}}_t^\mathrm{T}-\frac{1}{n_s}\mathbf{\bar{F}}_s\mathbf{\bar{F}}_s^\mathrm{T})\mathbf{\bar{F}}_t,
    \label{eq:derivative}
\end{equation}
is equivalent to the following cubic equation:
\begin{equation}
    a\eta^3 + b\eta^2 + c\eta + d = 0,
    \label{eq:cubic_equation}
\end{equation}
with the coefficients being:
\begin{equation} 
\small
    a =  \frac{2\lambda}{n_c^2}\mathrm{tr}[\mathbf{D_2}\mathbf{D_2}],~b =  -\frac{6\lambda}{n_c^2}\mathrm{tr}[\mathbf{D}_\mathbf{F}\mathbf{D_2}],~d = -\frac{1}{2}\mathrm{tr}[\mathbf{D_2}],
    \label{eq:coefficient_a}
\end{equation}
\begin{equation} 
\small
    c = \mathrm{tr}[\mathbf{D_2}] +  \frac{2\lambda}{n_c}\mathrm{tr}[\mathbf{D_2}\mathbf{S}] + \frac{2\lambda}{n_c^2}\big(\mathrm{tr}[\mathbf{D}_\mathbf{F}\mathbf{D}_\mathbf{F}] + \mathrm{tr}[\mathbf{D}_\mathbf{F}\mathbf{D}_\mathbf{F}^\mathrm{T}] \big),
    \label{eq:coefficient_c}
\end{equation}
where $\mathbf{D_2} \equiv \mathbf{D}\mathbf{D}^\mathrm{T}$, $\mathbf{D}_\mathbf{F} \equiv \mathbf{D}\mathbf{\bar{F}}_t^\mathrm{T}$, $\mathbf{D}$ $\equiv$ $\frac{\mathrm{d}l}{\mathrm{d}\mathbf{\bar{F}}_t}$ and $\mathbf{S}$ $\equiv$ $\frac{1}{n_c}\mathbf{\bar{F}}_t\mathbf{\bar{F}}_t^\mathrm{T}-\frac{1}{n_s}\mathbf{\bar{F}}_s\mathbf{\bar{F}}_s^\mathrm{T}$.
\newline
\hrule
\begin{proof} 
First, we plug $\mathbf{\bar{F}}_t - \eta\frac{\mathrm{d}l}{\mathrm{d}\mathbf{\bar{F}}_t}$ $=$ $\mathbf{\bar{F}}_t - \eta\mathbf{D}$ into the loss function in Eq.~\ref{eq:loss_function} and we get: 
\begin{equation}
    \begin{aligned}
    l(\mathbf{\bar{F}}_t - \eta\mathbf{D}) =& \underbrace{||\mathbf{A}||^2_2}_{\text{content loss}} + \lambda \underbrace{||\mathbf{B}||^2_2}_{\text{style loss}} \\
    =&~ \mathrm{tr}[\mathbf{A}\mathbf{A}^\mathrm{T}] + \lambda\cdot \mathrm{tr}[\mathbf{B}\mathbf{B}^\mathrm{T}],
    \end{aligned}
\end{equation}
where:
\begin{equation}
\small
    \begin{aligned}
    \mathbf{A} =~&\mathbf{\bar{F}}_t - \mathbf{\bar{F}}_c - \eta\mathbf{D}, \\
    \mathbf{B} =~&\frac{1}{n_c}(\mathbf{\bar{F}}_t - \eta\mathbf{D})(\mathbf{\bar{F}}_t - \eta\mathbf{D})^\mathrm{T}-\frac{1}{n_s}\mathbf{\bar{F}}_s\mathbf{\bar{F}}_s^\mathrm{T} \\
    =~& (\frac{1}{n_c}\mathbf{\bar{F}}_t\mathbf{\bar{F}}_t^\mathrm{T}-\frac{1}{n_s}\mathbf{\bar{F}}_s\mathbf{\bar{F}}_s^\mathrm{T}) - \frac{1}{n_c}\big(\eta(\mathbf{D}\mathbf{\bar{F}}_t^\mathrm{T})^\mathrm{T}+\eta\mathbf{D}\mathbf{\bar{F}}_t^\mathrm{T}-\eta^2\mathbf{D}\mathbf{D}^\mathrm{T}\big) \\
    =~& \mathbf{S} - \frac{1}{n_c}\big(\eta(\mathbf{D}\mathbf{\bar{F}}_t^\mathrm{T})^\mathrm{T}+\eta\mathbf{D}\mathbf{\bar{F}}_t^\mathrm{T}-\eta^2\mathbf{D}\mathbf{D}^\mathrm{T}\big) \\
    =~& \mathbf{S} - \frac{1}{n_c}\left(\eta\mathbf{D_F^\mathrm{T}} + \eta\mathbf{D_F} - \eta^2\mathbf{D_2} \right)
    \end{aligned}
\end{equation}

To find the minimum, we differentiate $l(\mathbf{\bar{F}}_t - \eta\mathbf{D})$ with respect to $\eta$ and solve the derivative equal to zero. Before the differentiation, we first derive an useful the identity that given a matrix $\mathbf{M}$ the derivative $\frac{\mathrm{d}(\mathrm{tr}[\mathbf{M}\mathbf{M}^\mathrm{T}])}{\mathrm{d}p}$ with respect to a parameter $p$ is equal to $2\mathrm{tr}[\frac{\mathrm{d}\mathbf{M}}{\mathrm{d}p}\mathbf{M}^\mathrm{T}]$:
\begin{equation}
\begin{aligned}
    \frac{\mathrm{d}(\mathrm{tr}[\mathbf{M}\mathbf{M}^\mathrm{T}])}{\mathrm{d}p} 
    =~& \mathrm{tr}\big[\frac{\mathrm{d\mathbf{M}}}{\mathrm{d}p}\mathbf{M}^\mathrm{T}] + \mathrm{tr}[\mathbf{M}\frac{\mathrm{d\mathbf{M}^\mathrm{T}}}{\mathrm{d}p}\big] \\
    \overset{(*)}{=}~& \mathrm{tr}\big[\frac{\mathrm{d\mathbf{M}}}{\mathrm{d}p}\mathbf{M}^\mathrm{T}] + \mathrm{tr}[\big(\mathbf{M}\frac{\mathrm{d\mathbf{M}^\mathrm{T}}}{\mathrm{d}p}\big)^\mathrm{T}\big] \\
    =~& 2 \mathrm{tr}\big[\frac{\mathrm{d\mathbf{M}}}{\mathrm{d}p}\mathbf{M}^\mathrm{T}],
\end{aligned}
    \label{eq:derivative_MMT}
\end{equation}
where in (*) we use the identity that the trace of a matrix is equal to the trace of its transpose.
With the identity in Eq.~\ref{eq:derivative_MMT}, we have:
\begin{equation}
    \frac{\mathrm{d}l(\mathbf{\bar{F}}_t - \eta\mathbf{D})}{\mathrm{d}\eta} = \cancel{2}\mathrm{tr}\big[\frac{\mathrm{d}\mathbf{A}}{\mathrm{d}\eta}\mathbf{A}^\mathrm{T}\big] + \cancel{2}\lambda\cdot \mathrm{tr}\big[\frac{\mathrm{d}\mathbf{B}}{\mathrm{d}\eta}\mathbf{B}^\mathrm{T}\big] = 0,
    \label{eq:derivative_wrt_eta}
\end{equation}
where the $2$'s are crossed out and the equality to zero still holds. In detail, we have: 
\begin{equation}
    \begin{aligned}
        &\mathrm{tr}\big[\frac{\mathrm{d}\mathbf{A}}{\mathrm{d}\eta}\mathbf{A}^\mathrm{T}\big] \\
        =~& \mathrm{tr}[-\mathbf{D}(\mathbf{\bar{F}}_t - \mathbf{\bar{F}}_c - \eta\mathbf{D})^\mathrm{T}] \\
        =~& \mathrm{tr}[-\mathbf{D}(\mathbf{\bar{F}}_t - \mathbf{\bar{F}}_c)^\mathrm{T} + \eta\mathbf{D_2})]
    \end{aligned}
    \label{eq:derivative_1stterm}
\end{equation}
and:
\begin{equation}
\begin{aligned}
    \mathrm{tr}\big[\frac{\mathrm{d}\mathbf{B}}{\mathrm{d}\eta}\mathbf{B}^\mathrm{T}\big] 
    =& -\frac{1}{n_c}\mathrm{tr}\big[\left(\mathbf{D_F^\mathrm{T}} + \mathbf{D_F} - 2\eta\mathbf{D_2} \right)\mathbf{B}^\mathrm{T}\big] \\
    =& -\frac{1}{n_c}\mathrm{tr}\big[\left(\mathbf{D_F^\mathrm{T}} + \mathbf{D_F} - 2\eta\mathbf{D_2} \right)\mathbf{B}\big],
    \label{eq:_derivative_2ndterm}
\end{aligned}
\end{equation}
where the identity $\mathbf{B}$ $=$ $\mathbf{B}^\mathrm{T}$ is used. Due to the following equality:
\begin{equation}
\begin{aligned}
    \mathrm{tr}\big[(\mathbf{D_F^\mathrm{T}}\mathbf{B}\big] &\overset{(a)}{=}  \mathrm{tr}\big[\mathbf{B}^\mathrm{T}\mathbf{D_F}\big] \\
    & \overset{(b)}{=}\mathrm{tr}\big[\mathbf{B}\mathbf{D_F}\big] \\
    & \overset{(c)}{=}\mathrm{tr}\big[\mathbf{D_F}\mathbf{B}\big],
\end{aligned}
\end{equation}
where we use in (a) the identity $\mathrm{tr}[\mathbf{M}^\mathrm{T}]$ = $\mathrm{tr}[\mathbf{M}]$ for any square matrix $\mathbf{M}$, in (b) the identity $\mathbf{B}$ $=$ $\mathbf{B}^\mathrm{T}$, and in (c) the identity  $\mathrm{tr}[\mathbf{M}_1\mathbf{M}_2]$ = $\mathrm{tr}[\mathbf{M}_2\mathbf{M}_1]$ for any square matrix that can be decomposed into the product of two matrices $\mathbf{M}_1$ and $\mathbf{M}_2$, Eq.~\ref{eq:_derivative_2ndterm} can be further written as: 
\begin{equation}
\small
\begin{aligned}
    &\mathrm{tr}\big[\frac{\mathrm{d}\mathbf{B}}{\mathrm{d}\eta}\mathbf{B}^\mathrm{T}\big] \\
    =& -\frac{2}{n_c}\mathrm{tr}\big[\big(\mathbf{D_F}-\eta\mathbf{D_2}\big)\mathbf{B}\big]\\
    =&-\frac{2}{n_c}\mathrm{tr}\left[\big(\mathbf{D_F}-\eta\mathbf{D_2}\big)\left(\mathbf{S} - \frac{1}{n_c}\left(\eta\mathbf{D_F^\mathrm{T}} + \eta\mathbf{D_F} - \eta^2\mathbf{D_2} \right)\right)
    \right].
    \label{eq:derivative_2ndterm}
\end{aligned}
\end{equation}
If we substitute Eq.~\ref{eq:derivative_1stterm} and Eq.~\ref{eq:derivative_2ndterm} into Eq.~\ref{eq:derivative_wrt_eta} and group the terms by the order of $\eta$, arranging them into the format $a\eta^3+b\eta^2+c\eta+d$ $=$ $0$, we have the coefficients being:
\begin{equation} 
    a =  \frac{2\lambda}{n_c^2}\mathrm{tr}[\mathbf{D_2}\mathbf{D_2}], 
    \label{eq:coefficient_a_proof}
\end{equation}
\begin{equation}
    b = -\frac{2\lambda}{n_c^2}\mathrm{tr}\left[\mathbf{D_2}\mathbf{D_F^\mathrm{T}} + \mathbf{D_2}\mathbf{D_F} + \mathbf{D_F}\mathbf{D_2}  \right], 
    \label{eq:_coefficient_b_proof}
\end{equation}
\begin{equation} 
    \begin{aligned}
    c =~ &\mathrm{tr}[\mathbf{D_2}] +  \frac{2\lambda}{n_c}\mathrm{tr}[\mathbf{D_2}\mathbf{S}]
    + \frac{2\lambda}{n_c^2}\big(\mathrm{tr}[\mathbf{D_F}\mathbf{D_F}] + \mathrm{tr}[\mathbf{D_F}\mathbf{D_F^\mathbf{T}}] \big),
    \label{eq:coefficient_c_proof}
    \end{aligned}
\end{equation}
\begin{equation}
    d = -\mathrm{tr}[\mathbf{D}(\mathbf{\bar{F}}_t-\mathbf{\bar{F}}_c)^\mathrm{T}] -\frac{2\lambda}{n_c}\mathrm{tr}[\mathbf{D}\mathbf{\bar{F}}_t^\mathrm{T}\mathbf{S}].
    \label{eq:_coefficient_d_proof}
\end{equation}

Since $\mathrm{tr}[\mathbf{D_2}\mathbf{D_F^\mathrm{T}}]$ is equal to $\mathrm{tr}[\mathbf{D_F}\mathbf{D_2}]$ due to the aforementioned trace identity $\mathrm{tr}[\mathbf{M}]$ $=$ $\mathrm{tr}[\mathbf{M}^\mathrm{T}]$, and $\mathrm{tr}[\mathbf{D_F}\mathbf{D_2}]$ is equal to $\mathrm{tr}[\mathbf{D_2}\mathbf{D_F}]$ due to the aforementioned trace identity $\mathrm{tr}[\mathbf{M}_1\mathbf{M}_2]$ = $\mathrm{tr}[\mathbf{M}_2\mathbf{M}_1]$, Eq.~\ref{eq:_coefficient_b_proof} can be further simplified as:
\begin{equation}
    b = -\frac{6\lambda}{n_c^2}\mathrm{tr}\big[\mathbf{D_F}\mathbf{D_2}\big].
    \label{eq:coefficient_b_proof}
\end{equation}
In addition, $d$ in Eq.~\ref{eq:_coefficient_d_proof} can be further simplified as follows:
\begin{equation}
    \begin{aligned}
    d &= -\mathrm{tr}[\mathbf{D}(\mathbf{\bar{F}}_t-\mathbf{\bar{F}}_c)^\mathrm{T}] -\frac{2\lambda}{n_c}\mathrm{tr}[\mathbf{D}\mathbf{\bar{F}}_t^\mathrm{T}\mathbf{S}]\\
    &= -\mathrm{tr}\big[\mathbf{D}\big((\mathbf{\bar{F}}_t-\mathbf{\bar{F}}_c)^\mathrm{T}+\frac{2\lambda}{n_c}(\mathbf{S}^\mathrm{T}\mathbf{\bar{F}}_t)^\mathrm{T} \big)\big]\\
    &\overset{(a)}{=} -\mathrm{tr}\big[\mathbf{D}\big((\mathbf{\bar{F}}_t-\mathbf{\bar{F}}_c)^\mathrm{T}+\frac{2\lambda}{n_c}(\mathbf{S}\mathbf{\bar{F}}_t)^\mathrm{T} \big) \big]\\
    &\overset{(b)}{=}-\frac{1}{2}\mathrm{tr}[\mathbf{D}\mathbf{D}^\mathrm{T}]\\
    &=-\frac{1}{2}\mathrm{tr}[\mathbf{D_2}],
    \end{aligned}
    \label{eq:coefficient_d_proof}
\end{equation}
where in (a) we use $\mathbf{S}^\mathrm{T}$ $=$ $\mathbf{S}$ and in (b) we introduce the equality in Eq.~\ref{eq:derivative}. We conclude the derivation with the Equations~(\ref{eq:coefficient_a_proof}), (\ref{eq:coefficient_b_proof}), (\ref{eq:coefficient_c_proof}), and (\ref{eq:coefficient_d_proof}).
\end{proof}

\section*{Proof of at least one positive solution to Equation 7 in the main paper}
To comply with the constraint $\eta$ $>$ $0$ in line search, we have to ensure that there is at least one positive solution to Equation 7 (Eq.~\ref{eq:cubic_equation} in this document). We prove this in the following.
\begin{proof}
For a cubic equation, there are two possible sets of solutions: either three real roots or one real root with two complex conjugated roots. From the relation between roots and coefficients, the product of three roots of Eq.~\ref{eq:cubic_equation} is equal to $-\frac{d}{a}$ $=$ $\frac{n_c^2}{4\lambda}\frac{\mathrm{tr}[\mathbf{D}\mathbf{D}^\mathrm{T}]}{\mathrm{tr}[\mathbf{D}\mathbf{D}^\mathrm{T}\mathbf{D}\mathbf{D}^\mathrm{T}]}$ $=$ $\frac{n_c^2}{4\lambda}\frac{||\mathbf{D}||^2_\mathrm{F}}{||\mathbf{D}\mathbf{D}^\mathrm{T}||^2_\mathrm{F}}$ $>$ $0$, where $||\cdot||_\mathrm{F}$ denotes the Frobenius norm. If the solutions to Eq.~\ref{eq:cubic_equation} are three real roots, since the product of three real roots is positive, at least one of them must be positive. If the solutions are one real root and two complex conjugated roots, since the product of three roots is positive and the product of two complex conjugated roots is the squared 2-norm of the conjugated roots, which is positive, the remaining real root should be positive. Therefore, there is at least one positive solution to Eq.~\ref{eq:cubic_equation}.
\end{proof}

\section*{Values of $\eta$ searched by LS-FT}
\begin{figure*}[h]
    \centering
    \includegraphics[width=\textwidth]{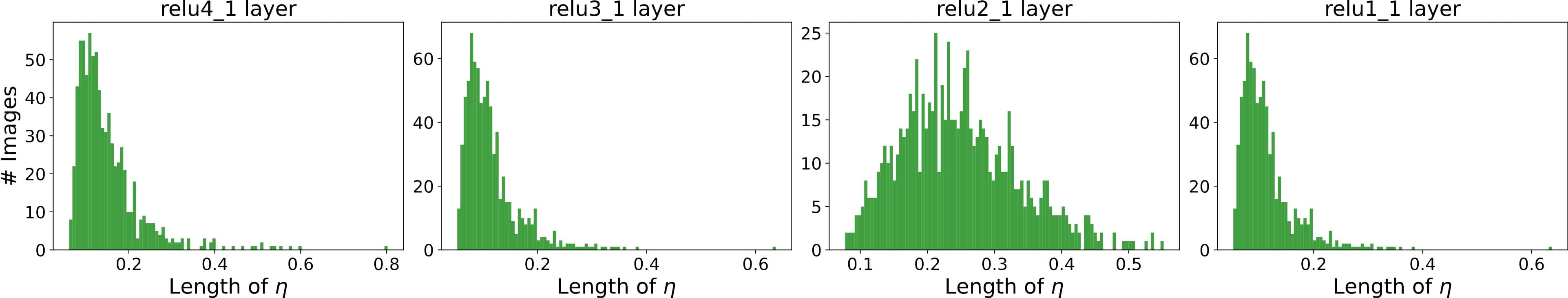}
    \caption{Histograms of the values of line-searched $\eta$ at the $\textit{reluN\_1}$ layers. This test is done using PhotoWCT$^2$~\cite{chiu2021photowct2} and the PST dataset~\cite{xia2020joint}.}
    \label{fig:eta_stats}
\end{figure*}

To illustrate how the values of line-searched $\eta$ differ from the constant $\eta$ equal to 0.01 used in IterFT~\cite{chiu2020iterative}, we embed LS-FT in PhotoWCT$^2$~\cite{chiu2021photowct2} and computed the values of line-searched $\eta$ at each $\textit{reluN\_1}$ layer using the PST dataset~\cite{xia2020joint}. As shown in Fig.~\ref{fig:eta_stats}, depending on the layer where LS-FT line-searches $\eta$ and the input images, the value of a line-searched $\eta$ can be as small as 0.04 and as large as 0.8.

\section*{Effect of the content-style control knob $\alpha$}

\begin{figure*}[t]
    \centering
    \includegraphics[width=\textwidth]{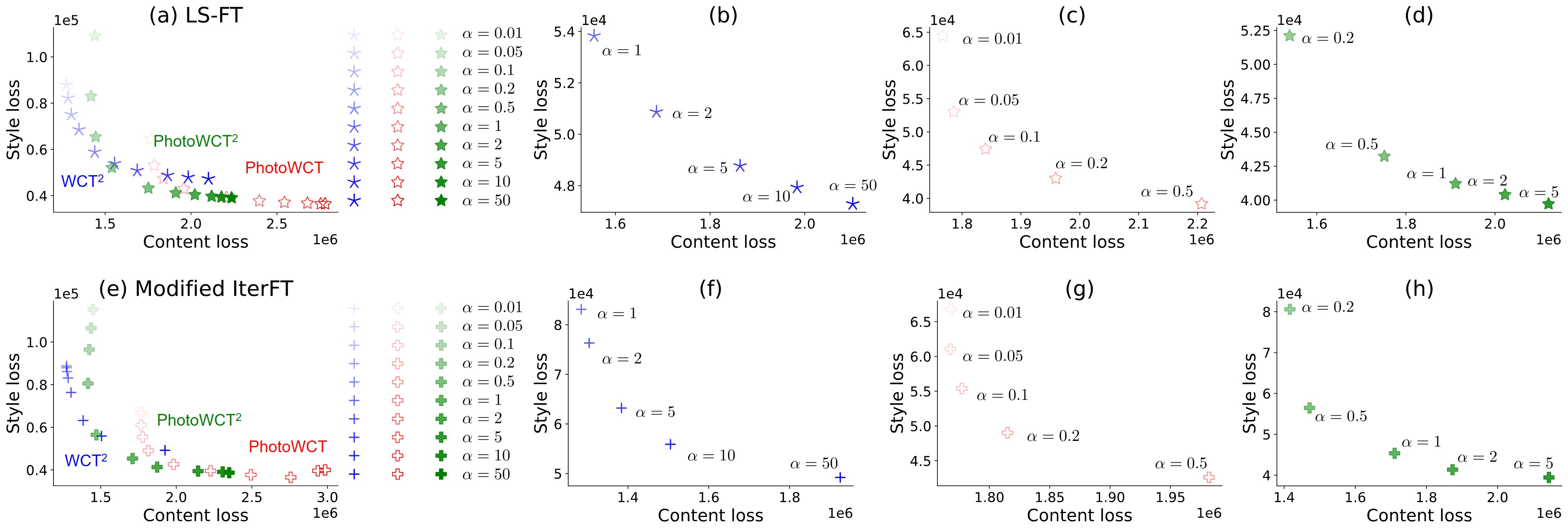}
    \caption{Effect of the content-style control knob $\alpha$ on WCT$^2$~\cite{yoo2019photorealistic}, PhotoWCT~\cite{li2018closed}, and PhotoWCT$^2$~\cite{chiu2021photowct2}.}
    \label{fig:effect_of_alpha}
    \vspace{-3mm}
\end{figure*}

We introduce a content-style control knob $\alpha$ in Section 3.3 in the main paper to adjust the value of $\lambda$ in the unit of $\frac{||\mathbf{\bar{F}}_c||^2_2}{||\frac{1}{n_s}\mathbf{\bar{F}}_s\mathbf{\bar{F}}_s^\mathrm{T}||^2_2}$ ($\lambda$ $=$ $\alpha\frac{||\mathbf{\bar{F}}_c||^2_2}{||\frac{1}{n_s}\mathbf{\bar{F}}_s\mathbf{\bar{F}}_s^\mathrm{T}||^2_2}$) such that we can tune the value of $\alpha$ to realize content-style controllability. As explained in Section 3.3 in the main paper, we want different values of $\alpha$ to boost the stylization strength of WCT$^2$~\cite{yoo2019photorealistic}, to boost the content preservation for PhotoWCT~\cite{li2018closed} and PCA-d~\cite{chiu2022pca}, and to balance content preservation and stylization strength for PhotoWCT$^2$~\cite{chiu2021photowct2}.

Fig.~\ref{fig:effect_of_alpha}(a) shows the performance of WCT$^2$, PhotoWCT, and PhotoWCT$^2$ with LS-FT of different $\alpha$ values, while Fig.~\ref{fig:effect_of_alpha}(e) shows the performance of Modified IterFT on the same models. The other panels in Fig.~\ref{fig:effect_of_alpha} are the zoomed plots for different models. To boost the stylization strength of WCT$^2$, we observe in Fig.~\ref{fig:effect_of_alpha}(b,f) that for LS-FT and Modified IterFT the style loss marginally decreases when $\alpha$ is larger than 10 and 50, respectively. Therefore, we set $\alpha$ equal to 10 and 50 for LS-FT and Modified IterFT, respectively, when they are used in the WCT$^2$ model.  To boost the content preservation ability of PhotoWCT, we observe in Fig.~\ref{fig:effect_of_alpha}(c,g) that for LS-FT and Modified IterFT the content loss marginally decreases when $\alpha$ is smaller than 0.2 and 0.5, respectively. Therefore, we set $\alpha$ equal to 0.2 and 0.5 for LS-FT and Modified IterFT, respectively, when they are used in the PhotoWCT model. To balance content preservation ability and stylization strength of PhotoWCT$^2$, we first recall in the Figure 5(d) in the main paper that the style loss of AdaIN is around 4.37e$^4$. Thus, to have stronger stylization strength than AdaIN and better content preservation ability than ZCA, we want the style losses of LS-FT and Modified IterFT smaller than 4.37e$^4$ and the content losses of LS-FT and Modified IterFT as small as possible. We observe in Fig.~\ref{fig:effect_of_alpha}(d,h) that $\alpha$ $=$ $1$ for LS-FT and $\alpha$ $=$ $2$ for Modified IterFT best fit this criterion, and so we set $\alpha$ as such values.

\begin{figure}
    \centering
    \includegraphics[width=\linewidth]{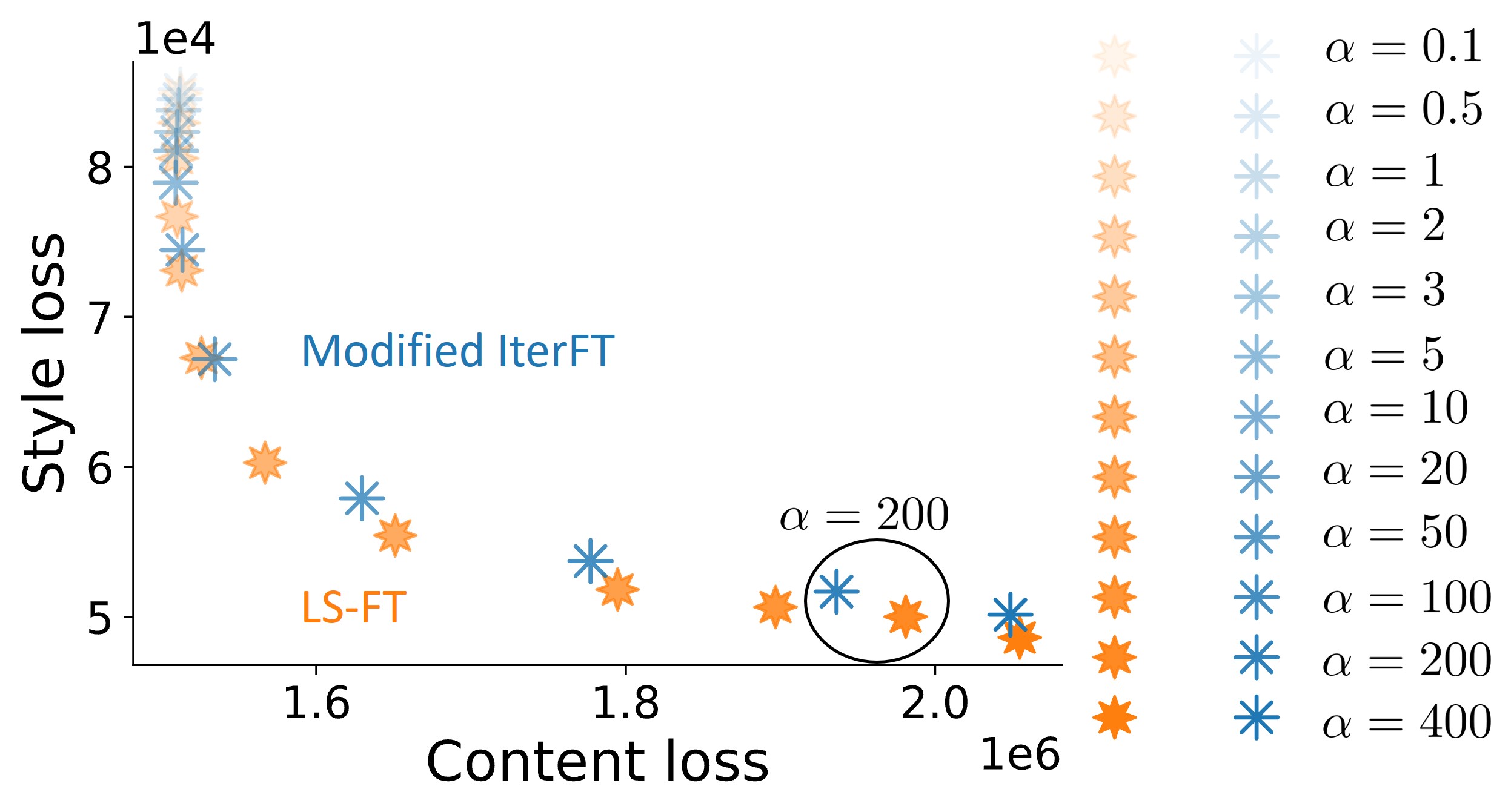}
    \caption{Effect of the content-style control knob $\alpha$ on PCA-d~\cite{chiu2022pca}.}
    \label{fig:effect_of_alpha_pca}
\end{figure}

Fig.~\ref{fig:effect_of_alpha_pca} shows the performance of PCA-d with LS-FT and Modified IterFT of different $\alpha$ values. We observe that for both LS-FT and Modified IterFT, when $\alpha$ is equal to 200, the content and style losses of PCA-d are very close to those from the aforementioned settings for WCT$^2$, PhotoWCT, and PhotoWCT$^2$. Therefore, we set $\alpha$ to 200 here.

\section*{Convergence comparison between Modified IterFT and LS-FT on WCT$^2$, PhotoWCT, and PCA-d}
\begin{figure*}[t]
    \centering
    \includegraphics[width=\textwidth]{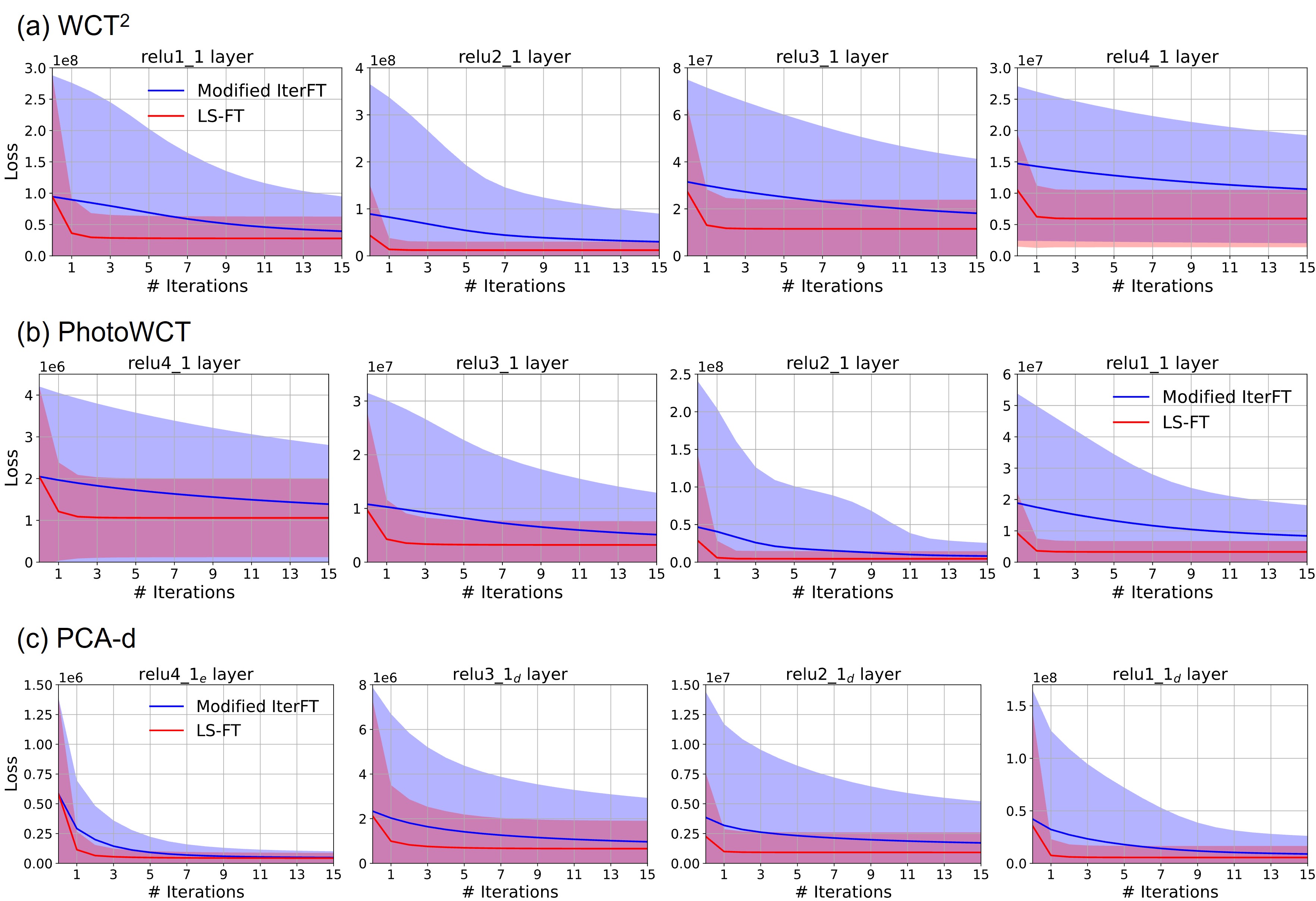}
    \caption{Convergence comparison between Modified IterFT and LS-FT on WCT$^2$~\cite{yoo2019photorealistic}, PhotoWCT~\cite{li2018closed}, and PCA-d~\cite{chiu2022pca}. The result shows that LS-FT needs only one iteration to outperform Modified IterFT at each transformation layer of each model, which is the same as the case of PhotoWCT$^2$~\cite{chiu2021photowct2}.}
    \label{fig:convergence_wct2_phwct}
    \vspace{-3mm}
\end{figure*}

We show in Figure 4 in the main paper the convergence comparison between Modified IterFT and LS-FT on PhotoWCT$^2$~\cite{chiu2021photowct2}. For completeness, we show in Fig.~\ref{fig:convergence_wct2_phwct} the convergence comparison between Modified IterFT and LS-FT on WCT$^2$~\cite{yoo2019photorealistic}, PhotoWCT~\cite{li2018closed}, and PCA-d~\cite{chiu2022pca}. The result shows that LS-FT needs only one iteration at each transformation layer of each model to outperform Modified IterFT, which is the same as the result from PhotoWCT$^2$.

\clearpage

\begin{table}[!t]
\setlength{\tabcolsep}{2.0pt}
    \centering
    \fontsize{8}{10}\selectfont
    \begin{tabular}{c c c c c c c c}
    \toprule
    {} & \multirow{2}{*}{PCA-d} & \multicolumn{4}{c}{Not Tunable} & \multicolumn{2}{c}{Tunable} \\
    \cmidrule(lr){3-6}\cmidrule(lr){7-8}
     &  & ZCA & OST & AdaIN & MAST & M-IterFT & LS-FT \\ \midrule
    HD & 0.024 & 0.032 & 0.048 & \bf{0.005} & 0.190 & 0.052 & \bf{0.031} \\
    FHD & 0.042 & 0.036 & 0.051 & \bf{0.006} & 0.211 & 0.093 & \bf{0.032} \\
    QHD & 0.070 & 0.064 & 0.072 & \bf{0.013} & 0.368 & 0.170 & \bf{0.064} \\
    UHD & 0.160 & 0.108 & 0.111 & \bf{0.027} & 0.572 & 0.374 & \bf{0.120} \\
    \bottomrule
    \end{tabular}
    \vspace{-0.25em}
    \caption{\small The speeds for stylization of images of PCA-d using different transformations. Unit: Second.}
    \label{tab:speed_pcad}
\end{table}

\section*{Unreasonable results from IterFT are fixed with our Modified IterFT and LS-FT}

\begin{figure*}[!t]
    \centering
    \includegraphics[width=\textwidth]{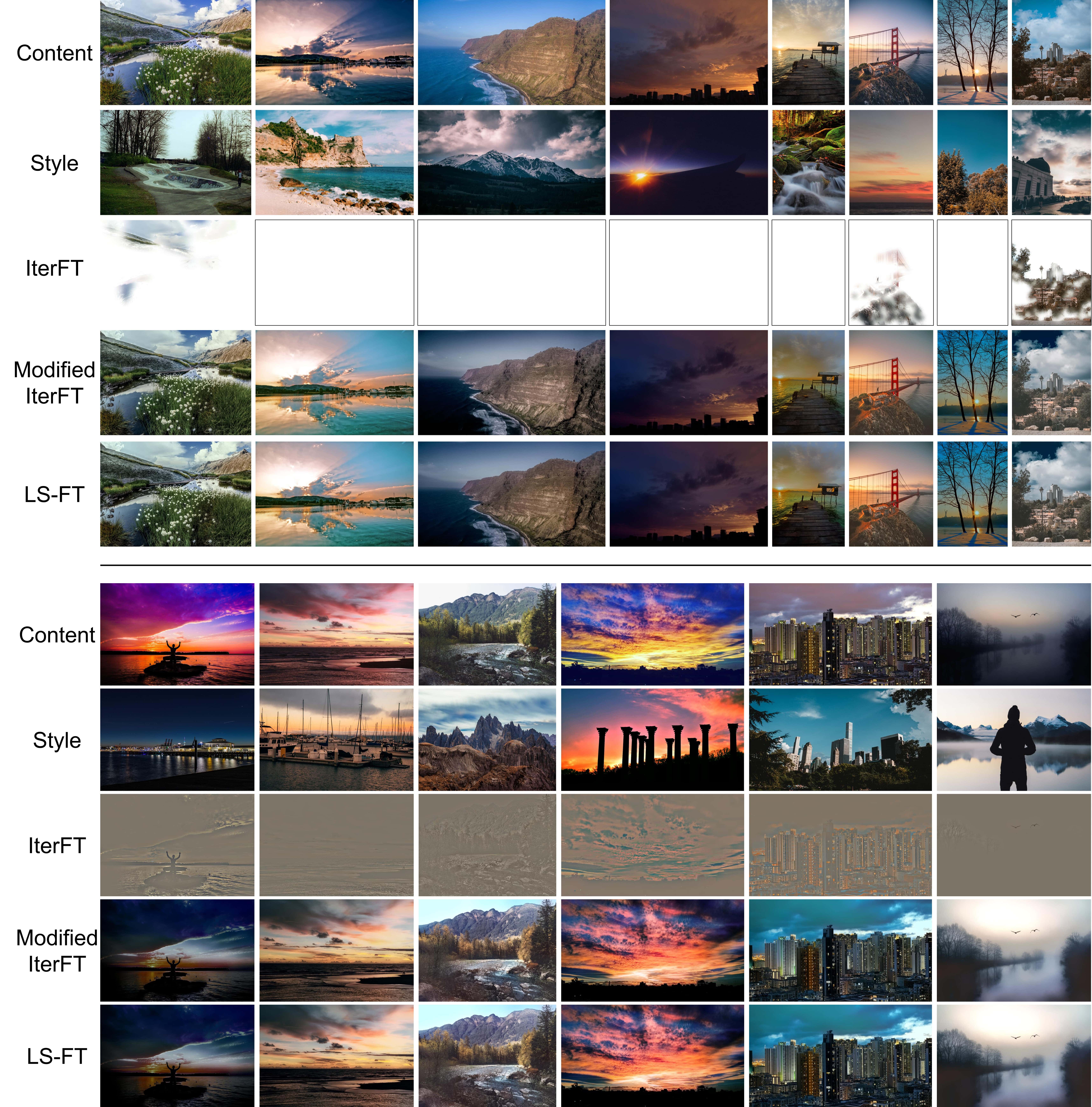}
    \caption{The failures of IterFT~\cite{chiu2020iterative} can be corrected with our Modified IterFT and LS-FT.}
    \label{fig:correction_IterFT}
\end{figure*}

We exclude IterFT~\cite{chiu2020iterative} from the experiment in Section 4.2 in the main paper, since it results in dozens of unreasonable results from the PST dataset~\cite{xia2020joint}. Here we show some failures from IterFT in Fig.~\ref{fig:correction_IterFT} and 
how they are corrected with our Modified IterFT and LS-FT. 

\section*{Speed of transformations on PCA-d}
We show the speed of different transformations on WCT$^2$, PhotoWCT, and PhotoWCT$^2$ in Table 2 in the main paper. Here we show the speed of transformations on PCA-d in Table~\ref{tab:speed_pcad}. We observe a similar result: our LS-FT is faster or comparably fast to ZCA.

\clearpage

{\small
\bibliographystyle{ieee_fullname}
\bibliography{egbib}
}

\end{document}